\pdfoutput=1
\documentclass[11pt]{article}

\usepackage[preprint]{acl}

\usepackage{times}
\usepackage{latexsym}
\usepackage{dblfloatfix}

\usepackage[T1]{fontenc}
\usepackage[utf8]{inputenc}

\usepackage{microtype}

\usepackage{inconsolata}

\usepackage[noend]{algorithmic}
\usepackage{amsfonts}       % blackboard math symbols
\usepackage{amsmath,amssymb}        % expand math equations
\usepackage{arydshln}
\usepackage[toc,page,header]{appendix}

\usepackage[utf8]{inputenc} % allow utf-8 input
\usepackage[T1]{fontenc}    % use 8-bit T1 fonts
\usepackage{hyperref}       % hyperlinks
\usepackage{url}            % simple URL typesetting
\usepackage{booktabs}       % professional-quality tables
\usepackage{latexsym}
\usepackage{listings}
\usepackage{nicefrac}       % compact symbols for 1/2, etc.
\usepackage{makecell}
\usepackage{microtype}      % microtypography
\usepackage{minitoc}
\mtcsetfeature{minitoc}{before}{\vspace{-2cm}}
\usepackage{multicol}
\usepackage{multirow}
\usepackage{tabularx}
\usepackage{float}
\usepackage{cleveref}
\usepackage{titletoc}
\usepackage{graphicx}
\usepackage{color, colortbl}
\usepackage{times}
\usepackage{wrapfig}
\usepackage{framed}
\usepackage{xcolor}         % colors
\usepackage{listings}
\definecolor{red-color}{RGB}{238, 117, 120}
\definecolor{green-color}{RGB}{112, 173, 71}
\definecolor{blue-color}{RGB}{91, 155, 213}
\definecolor{orange-color}{RGB}{237, 125, 49}
\definecolor{dkgreen}{rgb}{0,0.6,0}
\definecolor{LightGrey}{HTML}{d9d9d9}
\definecolor{mauve}{rgb}{0.58,0,0.82}

\title{\textit{Refiner}: Restructure Retrieval Content Efficiently to Advance Question-Answering Capabilities}

\author{
 \textbf{Zhonghao Li \textsuperscript{1}},
 \textbf{Xuming Hu\textsuperscript{1}},
 \textbf{Aiwei Liu\textsuperscript{2}},
 \textbf{Kening Zheng\textsuperscript{1}},
\\
 \textbf{Sirui Huang\textsuperscript{3}},
 \textbf{Hui Xiong\textsuperscript{1}},
\\
 \textsuperscript{1}Hongkong University of Science and Technology,
 \textsuperscript{2}Tsinghua University,
\\
 \textsuperscript{3}The Hong Kong Polytechnic University,
\\
\href{mailto:zhonghaoli@hkust-gz.edu.cn}{zhonghaoli@hkust-gz.edu.cn},
\href{mailto:xuminghu@hkust-gz.edu.cn}{xuminghu@hkust-gz.edu.cn},
\href{mailto:xionghui@hkust-gz.edu.cn}{xionghui@hkust-gz.edu.cn}
}

\begin{document}
\doparttoc % Tell to minitoc to generate a toc for the parts
\faketableofcontents

\maketitle
\begin{abstract}

Large Language Models (LLMs) are limited by their parametric knowledge, leading to hallucinations in knowledge-extensive tasks. To address this, Retrieval-Augmented Generation (RAG) incorporates external document chunks to expand LLM knowledge. Furthermore, compressing information from document chunks through extraction or summarization can improve LLM performance. Nonetheless, LLMs still struggle to notice and utilize scattered key information, a problem known as the "lost-in-the-middle" syndrome. Therefore, we typically need to restructure the content for LLM to recognize the key information.
% However, integrating those processed contents without considering and maintaining their information distinction can result in distorted and disjoined content, posing comprehension challenges for LLMs.
We propose \textit{Refiner}, an end-to-end extract-and-restructure paradigm that operates in the post-retrieval process of RAG. \textit{Refiner} leverages a single decoder-only LLM to adaptively extract query-relevant contents verbatim along with the necessary context, and section them based on their interconnectedness, thereby highlights information distinction, and aligns downstream LLMs with the original context effectively.
Experiments show that a trained \textit{Refiner} (with 7B parameters) exhibits significant gain to downstream LLM in improving answer accuracy, and outperforms other state-of-the-art advanced RAG and concurrent compressing approaches in various single-hop and multi-hop QA tasks. Notably, \textit{Refiner} achieves a 80.5\% tokens reduction and a 1.6-7.0\% improvement margin in multi-hop tasks compared to the next best solution.
\textit{Refiner} is a plug-and-play solution that can be seamlessly integrated with RAG systems, facilitating its application across diverse open-source frameworks.

\end{abstract}

\section{Introduction}

\begin{figure*}
    \centering
    \includegraphics[width=0.95\linewidth]{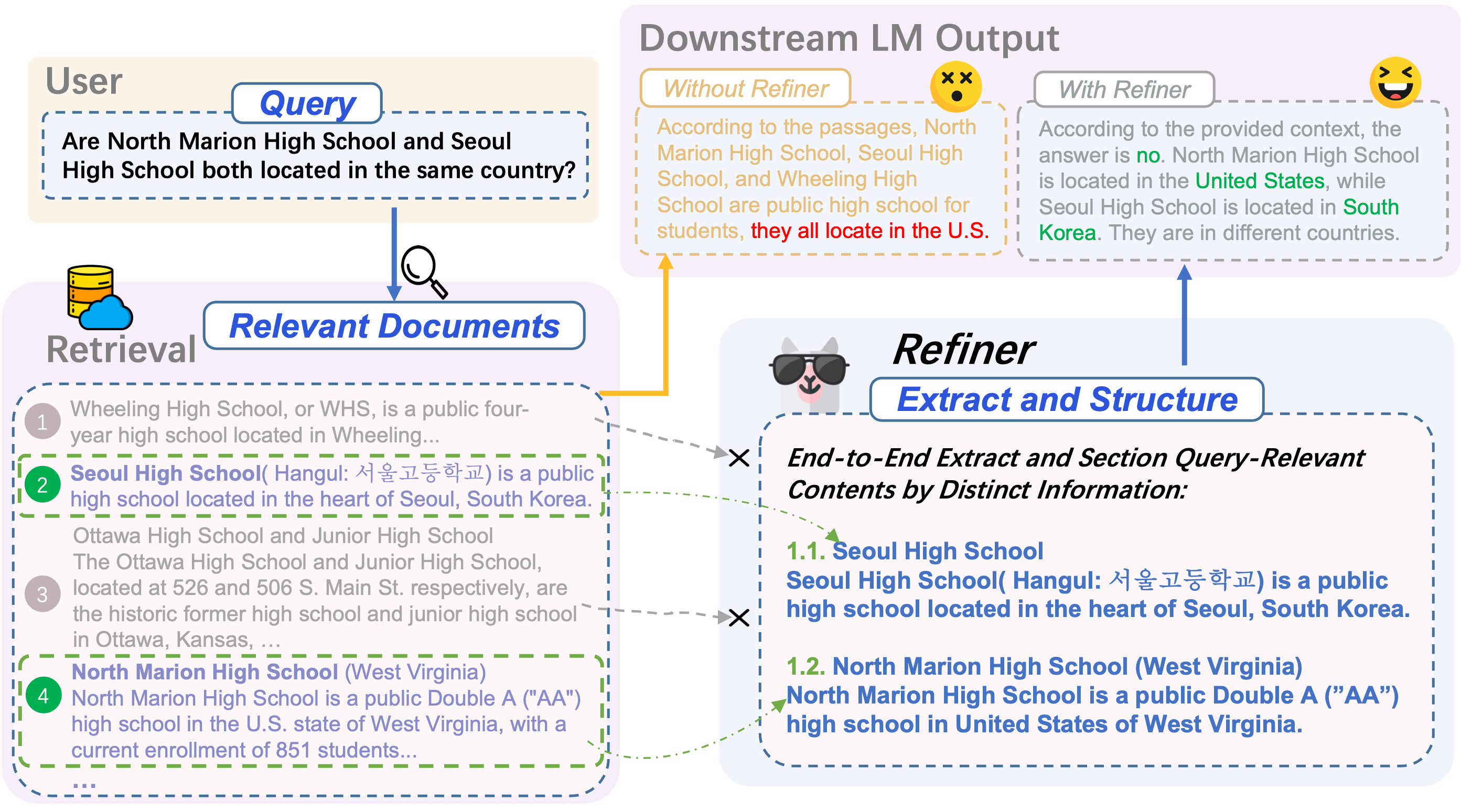}
    \caption{The example shows that misleading document chunks are prone to impeding downstream LM from comprehension and leading to hallucination. While our \textit{Refiner} mitigates the problem by extracting and structuring query-relevant contents by sectioning distinct information, making downstream LM disseminate them effortlessly.}
    \label{fig:how_refiner_works}
\end{figure*}

Recent Large Language Models (LLMs)~\citep{openai2024gpt4} have demonstrated their remarkable abilities in understanding query~\citep{brown2020language}~\citep{ouyang2022training} and generating human language texts.
Nonetheless, LLMs still face significant limitations,
particularly in domain-specific~\citep{zhang2024raft} or knowledge-intensive~\citep{kandpal2023large} tasks, where they often hallucinate~\citep{zhang2023sirens} when handling queries beyond their parametric knowledge~\citep{muhlgay2024generating}.
To address this issue, Retrieval-augmented generation (RAG)~\citep{lewis2021retrievalaugmented} has been rapidly developed~\citep{gao2024retrievalaugmented}, leveraging external knowledge base to retrieve document chunks semantically similar to the query and incorporate them into LLMs, thereby generating more faithful~\citep{khandelwal2020generalization} and generalizable~\citep{kamalloo2023evaluating} outputs.

However, generating answers from the top retrieved document chunks~\citep{karpukhin2020dense} indiscriminately has various shortcomings.
Due to the potential ambiguity or insufficiency of information in the query, the chunks could become contradictory and distracting, leading to downstream LLMs~\citep{izacard2022atlas} to hallucinate~\citep{zhang2023sirens}~\citep{Ji_2023}.
To mitigate this issue, researchers have proposed advanced RAG approaches, such as adopting query rewriting mechanism~\citep{chan2024rqrag} or self-reflection tokens~\citep{asai2023selfrag} during the generation process, which enhances the performance of RAG by enabling on-demand retrieval and selecting optimal answer generated from each document chunk. Meanwhile, others focus on compressing input prompt by employing a language model (LM)~\citep{anderson-etal-2022-lingua}~\citep{yang-etal-2023-prca}~\citep{jiang-etal-2023-longllmlingua} to exclude irrelevant content, thereby reduces computational costs and alleviate "lost-in-the-middle"~\citep{wang2023learning} syndrome.

Despite these advancements, LMs are still predisposed to neglect key information scattered across multiple chunks. Moreover, the document chunks, retrieved based on semantic proximity to a query, often pertain to identical subjects and occasionally contain similar yet distinguished information. Failing to maintain context and naively concatenating these text chunks, can induce confusion and biased comprehension in downstream LM (See examples in~\autoref{sec: cases-in-single-hop-eval-dataset}), exacerbating "lost-in-the-middle" syndrome. This is particularly significant in multi-hop question answering (QA)\citep{yang2018hotpotqa}~\citep{xanh2020_2wikimultihop} task, which requires LM to synthesize information from multiple contents to arrive at a correct answer. Therefore, we typically need to structure the information for downstream LM to differentiate the key information.

We propose \textit{Refiner}, a novel end-to-end extract-and-restructure paradigm that takes into account the relatedness of information across document chunks. The proposal is realized from two approaches: (1). In order for downstream LMs to align with the original context, we propose to keep query-relevant contents verbatim, while also maintain context around those contents on-demand.
(2). To emphasize information relatedness among document chunks, we propose to structure the output, as illustrated in the bottom right of \autoref{fig:how_refiner_works}: The \textit{Refiner} servers as grouping extractive contents by different information using hierarchical section, wherein contents with similar information are grouped in the same section, while contents with disparate information are segregated into separate sections. This approach aims at further clarifying chunk-level information relatedness for downstream LMs to easily comprehend. A case study of how \textit{Refiner} resolves the relatedness information in practice, is illustrated in \autoref{tab:popqa-example1}.

The \textit{Refiner} leverages an arbitrary pre-trained LLM to generate targeted content, on the assumption that LLM possesses a broad foundation of world knowledge~\citep{zhou2024grasping}~\citep{li2024recall}, enabling it to effectively section the intricate, nuanced content-wise relationships. The \textit{Refiner} is trained with knowledge distillation technique~\citep{jung2024impossible} to adaptively determine the amount of context to retain around key contents, and output with the proposed format.

Empirical results on several QA tasks, including single-hop and multi-hop tasks, demonstrate that \textit{Refiner} significantly improves the answer accuracy of downstream LMs, surpassing the previous state-of-the-art (SOTA) advanced RAG solution and concurrent prompt compressing work by a margin of 2.2\%-7.0\% on two multi-hop QA datasets, and attains comparable accuracy on three single-hop QA datasets with a token reduction of 77.8\% on average, compared to the second-best solution.

The \textit{Refiner}’s plug-and-play nature make it ideal for API-based models without parameter access. Once trained, the \textit{Refiner} can be applied across different upstream retrieval systems and downstream LLMs. The structured output, containing sections, titles and contents, could be further disassembled, leading to a easier source document retrospection or citation integration for industry application.

\section{Related Work}

\paragraph{Retrieval Augmented Generation}
 The Retrieval Augmented Generation (RAG)~\citep{lewis2021retrievalaugmented}~\citep{gao2024retrievalaugmented} fundamentally enhances the capabilities of LLMs by retrieving pertinent documents from an external knowledge base, which is made possible through the calculation of semantic similarity between user's query and documents. Notably, the de facto approach is to segment documents into chunks and construct text representation embedding upon these chunks. By leveraging external knowledge sources like Wikipedia, RAG effectively mitigates the issue of generating factually inaccurate content, thereby greatly enhancing the reliability of generated text. RAG typically involves a retrieval system for query-relevant knowledge indexing from a knowledge base, and a LM that serves as question answering interface.

\paragraph{Advanced RAG}
Advanced RAG model incorporates multiple optimization strategies that span both pre-retrieval and post-retrieval processes. For example, Self-RAG~\citep{asai2023selfrag} introduces a selective knowledge retrieval mechanism, which utilizes a retrieve token to determine the necessity of retrieval and employs multiple critic tokens to assess answer quality. Building upon this, RQ-RAG~\citep{chan2024rqrag} is trained to dynamically improve query by integrating rewriting, decomposition, and disambiguation functionalities. Furthermore, chain-of-thought (COT)~\citep{wang2024rat} and Tree-of-thought (TOT)~\citep{yao2023tree} are incorporated to RAG to tackle certain logical-demanding tasks, which proved to have significantly enhanced the performance of LLMs.

\paragraph{Concurrent Post-Retrieval Work}
Previous research has explored the application of language model-based approaches to further distill or extract query-relevant key information from retrieved document segments~\citep{lewis2021retrievalaugmented}.
Some studies have developed summarization models~\citep{lennox2023retrieveclustersummarize}~\citep{jung2024impossible} to derive crucial query-relevant information from retrieved document blocks, RECOMP~\citep{xu2023recomp} has designed a dual-encoder extractive compressor and a encoder-decoder abstractive model to expunge irrelevant context and summarize upon the remainder. Additionally, LongLLMLingua~\citep{jiang-etal-2023-longllmlingua} proposes a question-aware compression method with a configurable compression rate, allowing them to reduce computational cost and latency, while LLMLingua-2~\citep{wu2024llmlingua2} formulates prompt compression as a token classification problem, thereby retains the faithfulness of the compressed prompt.

Compared with recent studies~\citep{xu2023recomp}~\citep{yoran2024making}~\citep{jiang-etal-2023-longllmlingua} that are similar to our \textit{Refiner}, major differences are worth mentioning:
These methodologies aim at effective and efficient summarization/extraction without considering the semantic relatedness among them, while we dedicate to retain relatedness information by explicitly sectioning them to form a more structured content. Furthermore, we take contextual completeness into account for downstream LMs to align with the original context effortlessly.

\section{Problem Formulation}

Following previous work~\citep{xu2023recomp}, we use $\mathcal{X}=\left \{ q_{1},\mathcal{D}_{1},....,q_{\mathsf{i}},\mathcal{D}_{\mathsf{i}} \right \}$ to represent \textit{Refiner}'s input sequence set,
where $q_{\mathsf{i}}$ denotes query and $\mathcal{D}_{\mathsf{i}}=\left \{ d^1_\mathsf{i},....,d^\mathcal{K}_\mathsf{i}   \right \} $ denotes the corresponding top-$\mathcal{K}$ retrieved documents at step $i$.
The \textit{Refiner} is designated to extract from $\mathcal{D}_\mathsf{i}$ with respect to $q_{\mathsf{i}}$ to generate output sequence $y_\mathsf{i}$.
 
The output sequences of the model $y_\mathsf{i}$ should satisfy the following principles:
(1) \textbf{Concise}: The output should be verbatim and context-completed w.r.t retrieved documents, such that $y_\mathsf{i} \subseteq \mathcal{D_\mathsf{i}}$.
(2) \textbf{Sectioned}: contents extracted from different document chunks should all be numeric sectioned. Contextually related contents should be grouped into the same section. Conversely, contents with different information should be grouped to different sections.
(3) \textbf{Succinct}: The output length should be significantly shorter than total length of the retrieved documents $\mathcal{D}_\mathsf{i}$, such that $\left\lvert y_\mathsf{i} \right\rvert \ll \left\lvert D_\mathsf{i} \right\rvert$. All contents should be helpful to response to the query. If none of the documents $\mathsf{d}$ is relevant, then $\left\lvert y_\mathsf{i} \right\rvert = 0$.

\section{Proposed Method: \textit{Refiner}}

\begin{figure*}
    \centering
    \includegraphics[width=0.9\linewidth]{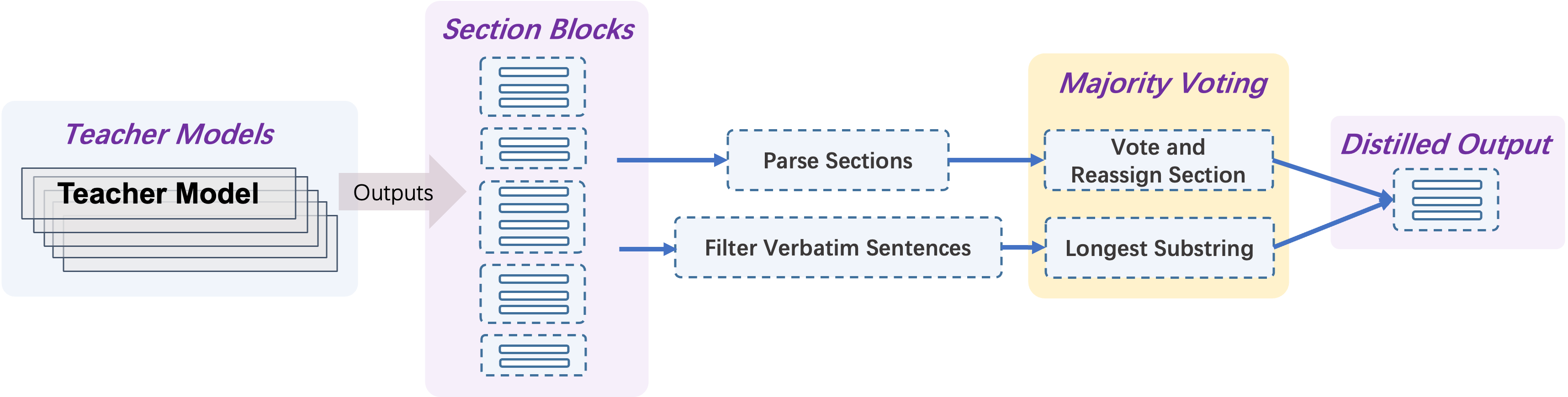}
    \caption{Knowledge distillation for \textit{Refiner}. The curated training data is generated by multiple teacher models, which undergoes parsing, filtering, voting to improve section and context expediency.}
    \label{fig:data-preparation}
\end{figure*}

The primary objective of our research is to develop a model that leverages intrinsic knowledge to structure content relevant to the question in the post-retrieval process. Empirical study~\citep{hu2024finetuning} shows that models with larger parameters tend to better conform their output format to instructions and exhibit enhanced capabilities in comprehension. However, larger models are accompanied by a quadratic increase in computational cost proportional to the length of the prompt. In our case, where the model is required to respond with brief content from a discursive, lengthy context, the computational cost necessitates particular attention. Therefore, we propose to fine-tune a smaller model in order to reign the computational costs and enhance its capabilities to conform to our proposed structured output format.

\subsection{Knowledge Distillation}
Accurate extraction of query-relevant content with pertinent context is crucial for the \textit{Refiner} model. To achieve this, we develop a training dataset that teaches \textit{Refiner} to determine the optimal scope of context. While a simple approach to maintain context might involve selecting a few sentences surrounding the key content from the original document chunk, this risks introducing irrelevant context. Instead, we rely on larger LLMs to adaptively determine the extent of context to retain. Specifically, we employ knowledge distillation ~\citep{jung2024impossible} technique to train \textit{Refiner}, where a simpler student model approximates the behavior of more intricate teacher
models~\citep{zuchniak2023multiteacher}.

However, single teacher model can be imperfect, occasionally producing unstructured output, forgetting to provide context or grouping with erroneous section. To address this, we utilize multi-teacher knowledge distillation~\citep{zuchniak2023multiteacher}, where multiple teacher models generate outputs from the same input, and the optimal output segment is selected to create a curated dataset.

As depicted in \autoref{fig:data-preparation}, we utilize five teacher models, which are prompted to section context-completed query-relevant content from provided document chunks. The generated sentences are rigorously verified against the original document chunks to ensure verbatimness. The section parts are voted on and rearranged, and only the sentences generated by a majority of the models are considered relevant context. The \textit{Refiner} is subsequently trained on concatenated sentences with section restoration. We design diverse prompts on various QA datasets for teacher models to consistently provide the proposed extraction. See detailed data preparation procedure in \autoref{alg:train-process}, composition of teacher model and hyperparameters in \autoref{tab:teacher-models}, and task-specific prompt templates in \autoref{tab:teacher_model_prompts}.

\subsection{Model Training}

Now that we have constructed the training dataset, we use Supervised Fine-Tuning (SFT)~\citep{sun2024supervised} to train a pre-trained Llama-2-7B-Chat~\citep{touvron2023llama} model. In particular, given a curated high-quality train dataset $\mathcal{S_{SFT}}=\left \{ \mathcal{X}_{i}, D_\mathsf{i}, \mathcal{Y}_i  \right \} _{i=1}^{N} $, where $N$ denotes number of samples in the dataset, the goal is to gain a model parameters $\mathcal{M}_{\theta}$ with the following training objective function:

{
\small
\begin{equation}
\underset{\theta}{minimize} \mathcal{L} (\theta;\mathcal{S}_{SFT}) = -\mathbb{E}_{\mathcal{X},\mathcal{Y}\sim\mathcal{S}_{SFT} }[\log{\mathcal{M_{\theta}}}(\mathcal{Y}|\mathcal{X})]
\end{equation}
}

where $\mathcal{L}$ denotes the negative likelihood objective function to minimize, $\mathcal{M}_{\theta}$ represents the model parameters. The
expectation $\mathbb{E}_{\mathcal{X},\mathcal{Y}\sim\mathcal{S}_{SFT} }$ averages over our dataset. Note that negative likelihood is adopted to formulate the loss function. Following previous works~\citep{lewis2021retrievalaugmented}~\citep{asai2023selfrag}~\citep{yan2024corrective}~\citep{xu2023recomp}, we train \textit{Refiner} $\mathcal{M_{\theta}}$ to generate a textual sequence 
$\left \{ y_{\mathsf{i}} \right \} _{i=1}^{N}$ that extracts core information and context in $\mathcal{D}$ relevant to $q$ with fewer tokens than $\mathcal{D}$. To ensure our model can effectively handle training datasets of varying sizes, we employ a progressive training approach~\citep{wu2024continual} for the \textit{Refiner}.
More training details are in the \autoref{sec:training-details}.

\section{Experiments}

We assess the effectiveness of our paradigm on two distinct criteria: its potential to enhance the performance of downstream LMs and its relative efficacy compared to other RAG approaches. We integrate it with various downstream LMs and open-domain question answering datasets, building upon the methodologies established in prior researches~\citep{asai2023selfrag}~\citep{yan2024corrective}~\citep{xu2023recomp}.

\subsection{Evaluate Tasks and Datasets}
We evaluate our \textit{Refiner} and downstream LMs on a diverse open-domain QA tasks, including short-form QA, long-form QA and multi-hop QA tasks. Throughout these experiments, we conduct zero-shot evaluations, where the prompts only provide instructions on tasks and outputs in contrast to few-shot examples~\citep{wei2022finetuned}~\citep{sanh2022multitask}. Detailed experiment environments, including prompt templates for \textit{Refiner} and downstream LMs, are listed in the \autoref{sec:evaluation-details}.

Following previous work~\citep{asai2023selfrag}, we use accuracy as evaluation metric on all test dataset. Notably, the applied accuracy metric is reported to not always reflects the performance of the RAG system, as it looks up the existence of answer string in the generated output, omitting the semantic coherence which results in biased evaluation. In ARC Challenge task, we construct a regular expression-based calculation, which may influence the results compared with other reported scores from advanced RAG solutions. See \autoref{acc-for-arc_c} for our motivation and methodology for evaluating ARC Challenge task specifically.

 \textbf{Single-Hop QA Dataset} includes three datasets, PopQA~\citep{mallen-etal-2023-trust}, TriviaQA-unfiltered~\citep{joshi2017triviaqa} and Reasoning Challenge (ARC-Challenge)~\citep{Clark2018ThinkYH}. The RAG system needs
to answer arbitrary questions about factual knowledge for PopQA and TriviaQA, and multiple-choice questions task about scientific event that occurs in daily life for ARC Challenge. For PopQA, we utilize long-tail subset,
consisting of 1,399 rare entity queries with monthly Wikipedia page views of less than 100. The ARC Challenge consists of 1,172 QA samples.
For TriviaQA, we employ the established validation and test split methodology adopted in prior research~\citep{asai2023selfrag}~\citep{min-etal-2019-discrete}~\citep{guu2020realm}, using 11,313 test queries for evaluation. We employ Contriever-MS MARCO~\citep{izacard2021contriever} as our retriever and retrieve ten documents for each query by default.

\textbf{Multi-hop QA Dataset} includes HotpotQA-dev-distractor~\citep{yang2018hotpotqa} dataset with 7,405 samples, and 2Wiki-dev~\citep{xanh2020_2wikimultihop} dataset with 12,576 samples. In contrast to the single-hop datasets, where document chunks are retrieved using our default retriever, we directly utilize the relevant content provided by these datasets.

\subsection{Advanced RAG and Concurrent Work}
% \textbf{Baselines}: We evaluate downstream models augmented with default retrieval system (i.e., without adopting \textit{Refiner}) at test time, we collect generated output given the query appended with the top five retrieved document chunks using the same retriever as in our system.

We compare \textit{Refiner}-augmented LMs with other advanced RAG solutions, including Self-RAG~\citep{asai2023selfrag} and RQ-RAG~\citep{chan2024rqrag}. Additionally, we assess the performance of concurrent compressors, including RECOMP abstractive compressor~\citep{xu2023recomp} and LongLLMLingua~\citep{jiang-etal-2023-longllmlingua}. Hyperparameter settings are detailed in the \autoref{sec:hyperparams-for-advanced-rag} and \autoref{sec:hyperparams-for-concurrent-compressor}. To facilitate a fair comparison, we maintain consistency with the experimental frameworks, utilizing the same retriever framework and downstream LMs as those used in \textit{Refiner}.

\section{Results} \label{sec:Results}

\begin{table*}[tbhp]
    \vspace*{-6mm}

\footnotesize
\setlength{\tabcolsep}{4.5pt}
\begin{center}
\begin{tabular}{@{}lccccc@{}}
\toprule

 & \multicolumn{3}{c}{\textbf{Single-Hop}} & \multicolumn{2}{c}{\textbf{Multi-Hop}} \\
 & \multicolumn{1}{c}{\textbf{POPQA}} & \multicolumn{1}{c}{\textbf{TQA}} &  \multicolumn{1}{c}{\textbf{ARC-C}} & \multicolumn{1}{c}{\textbf{HOTPOT}} & \multicolumn{1}{c}{\textbf{2WIKI}} \\
 
\textbf{LM} & (acc) & (acc) & (acc) & (acc) & (acc) \\

\midrule

\multicolumn{6}{l}{\textit{\textbf{Proprietary LLM}}} \\
$\text{GPT-3.5-Turbo}*$ & 29.3 & 74.3 & 75.3 & 57.3 & 54.3 \\

% \midrule

% \multicolumn{6}{l}{\textit{\textbf{Baselines}}} \\
% $\text{Llama2}_\text{7B}*$ & 38.2 & 42.5 & 48.0 & 16.7 & 18.7 \\ 
% $\text{Llama2}_\text{13B}*$ & 45.7 & 47.0 & 26.0 & 21.7 & 22.9 \\ 
% $\text{Llama2-FT}_\text{7B}*$ & 48.7 & 57.3 & 65.8 & 37.5 & 32.3 \\ 

\midrule

\multicolumn{6}{l}{\textit{\textbf{Advanced RAG}} } \\ 

$\text{Self-RAG}_\text{7B}*$ & 54.9 & 66.4 & 47.3 & 41.6 & 39.1 \\ 
$\text{Self-RAG}_\text{13B}*$ & 55.8 & 69.3 & 53.1 & 47.5 & 45.7 \\ 
$\text{RQ-RAG}*$ & 57.1 & -  & 57.4 & 62.6 & 44.8 \\
% \textit{Self-CRAG} & 61.8 & - & 86.2 & 67.28 & - & - \\

\midrule

\multicolumn{6}{l}{\textit{\textbf{With Downstream LM: }}$\textbf{Llama2-Chat}_\textbf{7B}$} \\
\rowcolor{LightGrey}
\textit{No Compressor} & 55.4 & 67.5 & 50.5 & 60.7 & 57.9 \\  
% \rowcolor{LightGrey}
% $\text{RECOMP}_\text{abstractive}$ & 48.9 & 63.8 & - & 40.8 & 53.6  \\ 
$\text{RECOMP}_\text{abstractive}$ & 52.6 & 65.0 & 18.7 & 47.3 & 58.5 \\   
LongLLMLingua & \textbf{60.3} & \textbf{70.5} & 51.6 & 61.3 & 55.3 \\  
$\textit{Refiner}$ & 59.3 & 68.9 & \textbf{51.8} & \textbf{65.1} & \textbf{61.0} \\ 

\cdashline{1-6}[3pt/2pt]\noalign{\vskip 0.5ex}

\multicolumn{6}{l}{\textit{\textbf{With Downstream LM: }}$\textbf{Llama2-Chat}_\textbf{13B}$} \\
\rowcolor{LightGrey}
\textit{No Compressor} & 54.8 & 71.8 & 57.3 & 63.8 & 58.4 \\  
$\text{RECOMP}_\text{abstractive}$ & 53.4 & 65.8 & 49.2 & 49.2 & 58.7 \\
LongLLMLingua & 58.3 & \textbf{71.6} & \textbf{57.8} & 60.1 & 56.2 \\
\textit{Refiner} & \textbf{59.4} & 70.6 & 55.1 & \textbf{67.1} & \textbf{61.6} \\

\cdashline{1-6}[3pt/2pt]\noalign{\vskip 0.5ex}
\multicolumn{6}{l}{\textit{\textbf{With Downstream LM: }}$\textbf{Meta-Llama3-Instruct}_\textbf{8B}$} \\
\rowcolor{LightGrey}
\textit{No Compressor} & 64.2 & 74.4 & 77.1 & 70.4 & 67.3 \\
$\text{RECOMP}_\text{abstractive}$ & 50.1 & 63.3 & 48.0 & 46.2 & 56.1 \\
LongLLMLingua & 59.3 & \textbf{71.2} & \textbf{72.0} & 64.9 & 54.2 \\
\textit{Refiner} & \textbf{59.7} & 69.3 & 71.1 & \textbf{67.2} & \textbf{63.7} \\ 
% \midrule

% \multicolumn{6}{l}{\textit{\textbf{Ours}}} \\
% \rowcolor{LightGrey}
% \textit{Refiner}  & 58.1 & 67.9 & - & 74.6 & 65.7  \\

\bottomrule
\end{tabular} 
\caption{Accuracy on single-hop and multi-hop tasks. We report results of proprietary LLM, advanced RAG that works without \textit{Refiner}, compared to off-the-shelf Llama2-(7,13)B-Chat and Meta-Llama3-Instruct-8B augmented by \textit{Refiner}. \textbf{Bold} numbers indicate the best performance among compressors. * indicates results reported by concurrent work. - indicates the corresponding task is not reported by the original paper.}
\label{tab:refiner-vs-rag}
\end{center}\vspace{-0.6em}
\vspace{-0.1in}
\end{table*}

\subsection{\textit{Refiner} consistently outperforms other approaches on multi-hop QA task}
As shown in \autoref{tab:refiner-vs-rag}, our \textit{Refiner} significantly outperforms existing advanced RAG variants and concurrent compressors on multi-hop QA tasks, while demonstrating competitive performance on single-hop QA tasks. Compared to the next best solutions, our \textit{Refiner}-augmented system registers a performance margin of 7.0\% on HotpotQA task, and 6.4\% on 2WikiMultihop task. Furthermore, the \textit{Refiner}-augmented LMs bridges the performance gap between small-scale and extreme-scale LLM on TriviaQA and HotpotQA tasks, surpasses GPT-3.5-Turbo model on PopQA, HotpotQA and 2WikiMultihop task, highlighting the substantial contribution of our \textit{Refiner} component to RAG. As anticipated, the performance gains fall short on single-hop tasks compared to LongLLMLingua, revealing that the structured output that distinguishes information disparity produced by \textit{Refiner} has a less pronounced impact on single-hop tasks than on multi-hop tasks.

% \begin{table}[t]
%     \small
%     \centering
%     \resizebox{0.5 \textwidth}{!}{%
%     \begin{tabular}{@{}lcccccc@{}}
%     \toprule

%     & \multicolumn{3}{c}{\textbf{Single-Hop}} & \multicolumn{2}{c}{\textbf{Multi-Hop}} \\
%     & \multicolumn{1}{c}{\textbf{POPQA}} & \multicolumn{1}{c}{\textbf{TQA}} &  \multicolumn{1}{c}{\textbf{ARC-C}} & \multicolumn{1}{c}{\textbf{HOTPOT}} & \multicolumn{1}{c}{\textbf{2WIKI}} & \\
%     \textbf{Input} & (avg \#token) & (avg \#token) & (avg \#token) & (avg \#token) & (avg \#token) & (rate)\\
%     \midrule

%     \textit{Document Chunks} & 1,520.5 & 2,054.8 & 1,933.0 & 1,531.7 & 1,121.8 & - \\ 

%     \cdashline{1-7}[3pt/2pt]\noalign{\vskip 0.5ex}

%     $\text{RECOMP}_\text{abstractive}$ & 21.0 & 144.1 & 20.7 & 28.0 & 64.6 & 96.6\% \\   
%     LongLLMLingua & 764.6 & 1,049.0 & 1,022.2 & 757.9 & 578.6 & 48.9\% \\  
%     $\textit{Refiner}$ & 143.1 & 206.7 & 280.2 & 130.0 & 131.2 & 89.1\% \\ 
%     \bottomrule
%     \end{tabular}%
%     }
%     \caption{Average tokens and compression rates among tasks and compressors.}
%     \label{tab:compression-rate-vs-rag}
% \end{table}

\begin{table}[t]
    \small
    \centering
    \resizebox{0.5 \textwidth}{!}{
    \begin{tabular}{lcccc@{}}
    \toprule

    & \multicolumn{1}{c}{\text{RECOMP} } & \multicolumn{1}{c}{LongLLMLingua} & \multicolumn{1}{c}{\textit{Refiner}} \\
    \textbf{Task} & (avg \#token) & (avg \#token) & (avg \#token) \\

    \midrule

    \textbf{POPQA} & 21.0 & 764.6 & 143.1 \\
    \textbf{TQA} & 144.1 & 1,049.0 & 206.7 \\
    \textbf{ARC-C} & 20.7 & 1,022.2 & 280.2 \\
    \textbf{HOTPOT} & 28.0 & 757.9 & 130.0 \\
    \textbf{2WIKI} & 64.6 & 578.6 & 131.2 \\
    
    \toprule
    
    \multicolumn{1}{c}{\textbf{Compress Rate}} & 96.6\% & 48.9\% & 89.1\% \\
    
    \bottomrule
    
    \end{tabular}
    }
    \caption{Average number of output tokens and compress rates among tasks and compressors.}
    \label{tab:compression-rate-vs-rag}
    \vspace{-0.9mm}
\end{table}

\subsection{\textit{Refiner}'s output is highly utilized by LMs}
To quantify the efficiency among compressors, we compare their compression rates by dividing the average number of output tokens generated by the compressors by the average number of input tokens in the document chunks, then subtracting the value from one. (see \autoref{tab:compression-rate-vs-rag}). We use Llama2-7B Chat tokenizer throughout the experiment. The results reveal that the RECOMP abstractive compressor achieves the highest compression rate, at 96.6\%, followed by \textit{Refiner}, which compresses 89.1\% of input tokens. LongLLMLingua's compression rate, set as a hyperparameter to 50\%, results in an actual compression rate of approximately 48.9\%.

Our \textit{Refiner} achieves a comparable performance to LongLLMLingua in single-hop QA tasks, with providing only 22.2\% of the average number of tokens, and surpasses it in multi-hop QA tasks with 19.5\% of tokens, demonstrating high utilization of \textit{Refiner}'s output by downstream LMs, attributable to its structured and context-completed output.

% To explore downstream LM utilization on \textit{Refiner}'s output, we calculate the extent of exploitation by dividing the accuracy of the downstream LM by that of the oracle. As indicated in \autoref{tab:refiner-vs-rag} (bottom two rows) and \autoref{tab:top_n} (oracle versus others), the \textit{Refiner}-aided LMs exhibit a substantial exploitation of the information provided by the \textit{Refiner}. Even in the case of Llama2-7B-Chat model, the utilization reaches 96.9\% on PopQA, 99.1\% on TriviaQA, 87.3\% on HotpotQA and 93.5\% on 2WikiMultihop task. Remarkably, the utilization exceeds 100\% on TriviaQA with Llama2-13B-Chat. We conclude that this is attribute to the the indirect information being further exploited by the more powerful LM. See \autoref{sec: cases-in-single-hop-eval-dataset} for case study on how \textit{Refiner} helps to highlight critical information.

\subsection{\textit{Refiner}-augmented system is less susceptible to noisy and lengthy content}
Recent study~\citep{xu2023recomp} reveals that appending irrelevant documents can negatively impact the performance of RAG, our \textit{Refiner} has successfully helped downstream LMs to mitigate the limitation by significantly reducing prompt length while keeping the performance of downstream LMs. Notably, our experimental results show that even when five additional irrelevant document chunks are appended, the performance of downstream LMs remains steady. In fact, as illustrated in \autoref{tab:top_n}, the accuracy of downstream LMs in both single-hop and multi-hop tasks has actually experienced an upswing. This content length tolerance exhibited by \textit{Refiner}-augmented system could alleviate the lost-in-the-middle~\citep{wang2023learning} phenomenon, commonly observed in downstream LMs. Such tolerance can be attributed to our training data, which deliberately organizes input document chunks in random order, thereby forcing the model to search and extract key information from full context.

% \subsection{The compressor family exhibits more performance gain on weaker LMs}
% Notably, our findings indicate that the compressor exhibits a performance increase of up to 4.9\% when paired with the weaker language model, Llama2-Chat-7B (see bold numbers and gray rows in \autoref{tab:refiner-vs-rag}). Conversely, the presence of compressors leads to a 5.1\% decrease in accuracy with the novel model, Meta-Llama3-Instruct-8B. This disparity manifests a crucial essence: compressors only benefits when they supply knowledge not already captured by the parameters of the downstream LMs, otherwise they may degrade performance.

\begin{table*}[tbhp]
    \vspace*{-6mm}
\footnotesize
\setlength{\tabcolsep}{5pt}
\begin{center}
\begin{tabular}{@{}lccccc@{}}
\toprule

 & \multicolumn{3}{c}{\textbf{Single-Hop}} & \multicolumn{2}{c}{\textbf{Multi-Hop}} \\
 & \multicolumn{1}{c}{\textbf{POPQA}} & \multicolumn{1}{c}{\textbf{TQA}} & \multicolumn{1}{c}{\textbf{ARC-C}} & \multicolumn{1}{c}{\textbf{Hotpot}} & \multicolumn{1}{c}{\textbf{2Wiki}} \\
 
\textbf{LM} & (acc) & (acc) & (acc) & (acc) & (acc) \\
\midrule
\multicolumn{6}{l}{\textbf{\textit{Refiner} of top 5 document chunks}} \\
% \rowcolor{LightGrey}
% \textit{Oracle} & 58.1 & 67.9 & - & 46.7 & 34.6 \\

$\textit{Refiner}+\text{Llama2-Chat}_\text{7B}$ & 59.3 & 68.9 & 51.1 & 47.4 & 46.0 \\ 
$\textit{Refiner}+\text{Llama2-Chat}_\text{13B}$ & 59.0 & 70.5 & 54.2 & 49.7 & 45.7 \\ 
$\textit{Refiner}+\text{Llama2-Chat}_\text{70B}$ & 58.3 & 72.2 & 55.9 & 52.3 & 50.1 \\ 
$\textit{Refiner}+\text{Meta-Llama3-Instruct}_\text{8B}$ & 58.3 & 69.1 & 71.8 & 49.0 & 48.0 \\ 
$\textit{Refiner}+\text{Meta-Llama3-Instruct}_\text{70B}$ & 57.9 & 70.9 & 84.1 & 49.2 & 46.7 \\ 
\hline
$\textit{\textbf{STD}}$ & \textit{0.57} & \textit{1.36} & \textit{14.06} & \textit{1.77} & \textit{1.80} \\

\hline
\midrule

\multicolumn{6}{l}{\textbf{\textit{Refiner} of top 10 document chunks}} \\
% \rowcolor{LightGrey}
% \textit{Oracle} & 61.2 & 69.5 & - & 74.6 & 65.7 \\

$\textit{Refiner}+\text{Llama2-Chat}_\text{7B}$ & 59.3 & 68.9 & 51.8 & 65.1 & 61.0 \\ 
$\textit{Refiner}+\text{Llama2-Chat}_\text{13B}$ & 59.4 & 70.6 & 55.1 & 67.1 & 61.6 \\ 
$\textit{Refiner}+\text{Llama2-Chat}_\text{70B}$ & 60.8 & 72.1 & 57.0 & 69.8 & 68.3 \\ 
$\textit{Refiner}+\text{Meta-Llama3-Instruct}_\text{8B}$ & 59.7 & 69.3 & 71.1 & 67.2 & 63.7 \\ 
$\textit{Refiner}+\text{Meta-Llama3-Instruct}_\text{70B}$ & 59.8 & 71.1 & 84.7 & 68.6 & 61.8 \\ 
\hline
$\textit{\textbf{STD}}$ & \textit{0.61} & \textit{1.31} & \textit{13.75} & \textit{1.77} & \textit{2.98} \\

\bottomrule
\end{tabular} 
\caption{Results of various downstream LMs in Meta Llama family augmented by \textit{Refiner} with top 5 and top 10 retrieved document chunks. A low standard deviation of accuracy score across downstream LMs in-task, indicating that the \textit{Refiner} is agnostic to both the downstream LM and the upstream retrieval content length.}
\label{tab:top_n}
\end{center}\vspace{-0.6em}
\vspace{-0.1in}
\end{table*}

\subsection{The \textit{Refiner} is resilient to RAG system}
To conduct a thorough investigation of the effectiveness coherency of \textit{Refiner} within different RAG settings, we undertook evaluation tasks using a range of \textit{Refiner}-augmented LMs and varied top retrieval settings, as illustrated in \autoref{tab:top_n}.
% In the case of single-hop datasets, given that retrieved document chunks are sorted in descending order according to their semantic distance to the query, it is frequently sufficient to employ a top 5 retrieval approach to retrieve crucial information for downstream generation. As a consequence, the differences in accuracy between single-hop oracles on comparable tasks are relatively small. In contrast, multi-hop tasks, whose test sets comprised randomly ordered document chunks, exhibit a substantial loss of information when half of the chunks are omitted during the process.
Notwithstanding the variations in content lengths and downstream LMs, the in-task performance exhibits its stability. As shown in \autoref{tab:top_n}, the standard deviation of in-task accuracy remains relatively constant, demonstrating its ubiquitous effectiveness, with the exception of the ARC Challenge task. This is due to the retrieved content's limited assistance with challenging questions, which necessitates downstream LMs to utilize intrinsic knowledge for deduction and extrapolation, eventually highlights the superiority of more sophisticated models.

\section{Ablation Study and Analysis}
To assess the efficacy of the proposed structured output, an ablation study is implemented by eliminating the structural components from the \textit{Refiner}'s output, resulting in naively concatenating the extracted content without any structural organization.
As illustrated in \autoref{tab:ablation-in-task}, we evaluate the impact of the removed structure on downstream tasks by testing models equipped with the expunged outputs, This includes using Llama2-(7B, 13B, 70B)-Chat, Meta-Llama3-(8B, 70B)-Instruct models, with both the top 5 and top 10 retrieval configurations. More detailed table from which data in \autoref{tab:ablation-in-task} is derived, are presented in \autoref{tab:ablation_in_model_top5} and \autoref{tab:ablation_in_model_top10}. Furthermore, we have conducted various revisions of structure format, the format and corresponding performance statistics are attached in \autoref{sec:more-analysis-on-output-structure}.

\begin{table*}[tbhp]
    \vspace*{-6mm}
\footnotesize
\setlength{\tabcolsep}{4.5pt}
\begin{center}
\begin{tabular}{@{}lccccc@{}}
\toprule

 & \multicolumn{3}{c}{\textbf{Single-Hop}} & \multicolumn{2}{c}{\textbf{Multi-Hop}} \\
 & \multicolumn{1}{c}{\textbf{POPQA}} & \multicolumn{1}{c}{\textbf{TQA}} & \multicolumn{1}{c}{\textbf{ARC-C}} & \multicolumn{1}{c}{\textbf{Hotpot}} & \multicolumn{1}{c}{\textbf{2Wiki}} \\
 
\textbf{LM} & \multicolumn{5}{c}{(acc decay w/o. structure)} \\
\midrule
\multicolumn{6}{l}{\textbf{Top 5 document chunks}} \\
% \rowcolor{LightGrey}
% \textit{Oracle} & 58.1 & 67.9 & - & 46.7 & 34.6 \\

$\textit{Refiner}+\text{Llama2-Chat}_\text{7B}$ & 1.6 & 2.4 & -0.6 & 0.1 & -0.3 \\ 
$\textit{Refiner}+\text{Llama2-Chat}_\text{13B}$ & 0.5 & 1.1 & -0.2 & 0.2 & -1.5 \\ 
$\textit{Refiner}+\text{Llama2-Chat}_\text{70B}$ & 0.3 & 0.8 & -2.8 & 0.1 & -0.1 \\ 
$\textit{Refiner}+\text{Meta-Llama3-Instruct}_\text{8B}$ & 0.4 & 2.3 & -2.6 & -0.2 & 0.0 \\ 
$\textit{Refiner}+\text{Meta-Llama3-Instruct}_\text{70B}$ & 0.3 & 2.0 & -1.4 & 0.5 & -0.1 \\ 
\midrule
$\textit{\textbf{MEAN}}$ & \textit{0.6} & \textit{1.7} & \textit{-1.5} & \textit{0.1} & \textit{-0.3} \\

\hline
\midrule

\multicolumn{6}{l}{\textbf{Top 10 document chunks}} \\
% \rowcolor{LightGrey}
% \textit{Oracle} & 61.2 & 69.5 & - & 74.6 & 65.7 \\

$\textit{Refiner}+\text{Llama2-Chat}_\text{7B}$ & 0.9 & 2.7
& 1.4 & 1.9 & 3.4 \\ 
$\textit{Refiner}+\text{Llama2-Chat}_\text{13B}$ & 1.5 & 2.3 & 0.1 & 0.8 & 0.2 \\ 
$\textit{Refiner}+\text{Llama2-Chat}_\text{70B}$ & 2.6 & 1.2 & -2.8 & 0.8 & 0.2 \\ 
$\textit{Refiner}+\text{Meta-Llama3}_\text{Instruct8B}$ & 1.9 & 2.1 & -2.2 & -0.2 & 0.1 \\ 
$\textit{Refiner}+\text{Meta-Llama3}_\text{Instruct70B}$ & 1.3 & 1.7 & 0.3 & 0.3 & 0.0 \\ 
\midrule
$\textit{\textbf{MEAN}}$ & \textit{1.6} & \textit{2.0} & \textit{-0.6} & \textit{0.7} & \textit{0.8} \\

\bottomrule
\end{tabular} 
\caption{Ablation analysis on various downstream LMs in Meta Llama family augmented by \textit{Refiner} with top 5 and top 10 retrieved document chunks. Accuracy decay and task-level statistic are presented by removing section parts from \textit{Refiner}'s structured output.}
\label{tab:ablation-in-task}
\end{center}\vspace{-0.6em}
\vspace{-0.1in}
\end{table*}

\subsection{Task-Level Analysis}
Our analysis of the decay in accuracy resulting from the removal of sections reveals significant variations across tasks. As shown in \autoref{tab:ablation-in-task}, the mean accuracy decay calculated for the top 5 and top 10 retrieval blocks indicates that sections make a substantial contribution to accuracy in most tasks, with most accuracy decays being positive. One notable exception is the ARC Challenge task, which experiences a decline in accuracy when sections are added. Further investigation reveals that this phenomenon stems from the retrieval system's propensity to retrieve documents related to incorrect options, leading to confusion for \textit{Refiner} and a bias towards incorrect information. Encouragingly, the importance of sections varies significantly across tasks, with the overall trend remaining consistent between top retrieval settings.

\subsection{Model-Level Analysis}
Another phenomenon has caught our attention: all downstream LMs under the top 10 retrieval setting have experienced a greater decay in accuracy compared to those under the top 5 retrieval settings. Our explanation is that \textit{Refiner} tends to extract more content from the extra document chunks that are remotely correlated to the query, and separates them from those that are closely related using sections. Without these sections, downstream LMs struggle to disseminate crucial information from the lengthy contents, leading to a degraded accuracy. More detailed statistical data in the \autoref{tab:ablation-in-task}, are attached in the \autoref{sec:ablation-detail}.

\subsection{How Verbatim is \textit{Refiner}'s Output?}
\begin{figure}
    \centering
    \includegraphics[width=1.\linewidth]{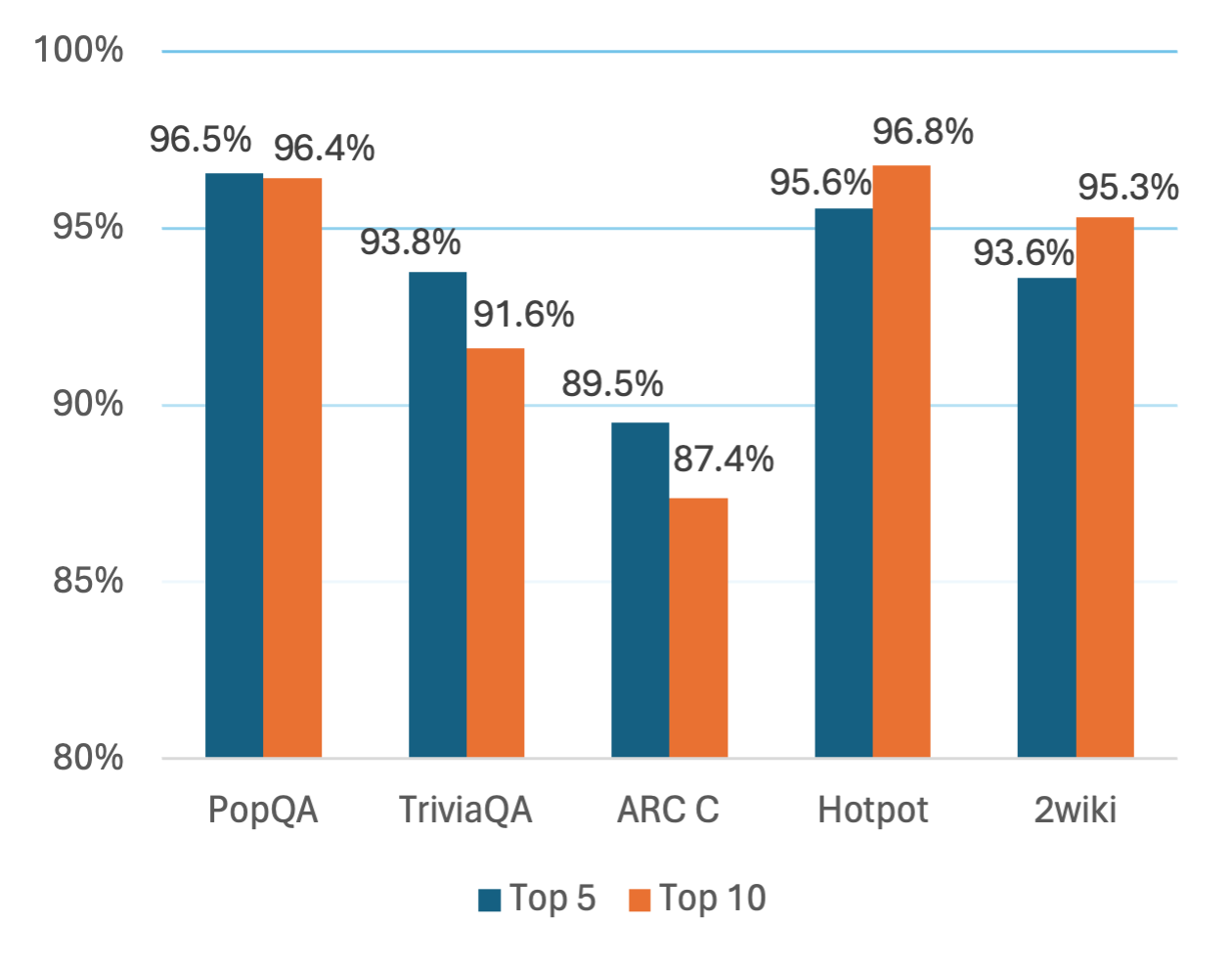}
    \caption{The ratio of contents extracted by \textit{Refiner} that are exactly originate from respective document chunks under various top retrieval settings and tasks.}
    \label{fig:ratio-of-verbatim-output}
\end{figure}

Like many other decoder-only models, \textit{Refiner} is a probabilistic LM~\citep{article} that sequentially generates the probability distribution of the next token given the preceding tokens. Although our objective is to train the model to produce output that selectively replicates the source content, there is always a possibility of non-conformity, wherein the model fails to reproduce the content verbatim from the given document chunks. The ratio of authentic output generated by \textit{Refiner} for each top retrieval setting and evaluation task dataset is calculated and illustrated in \autoref{fig:ratio-of-verbatim-output}. Overall, the model exhibits a range of conformity from 87.4\% to 96.8\%, whereupon we proceed to investigate its malformed output and discuss the underlying causes of this phenomenon.

\paragraph{Temptation to Answering}
The major predicament that \textit{Refiner} encounters, is its compulsion to answer to the query rather than extracting relevant information. This is often exacerbated when the query lacks challenge and the information in the document chunk directly answers to the query. Under such circumstances, the model is inclined to repeat the query and the information, rather than engaging in meaningful information extraction.

\paragraph{Temptation to Extract from Query}
Additionally, \textit{Refiner} is also prone to extracting informative content from the query, a problem that is particularly evident in the ARC-Challenge task. This task is characterized by a query comprising a question and four or more potential answers. The model is more inclined to extract content from these options, which, although potentially incorrect, are nonetheless relevant to the question at hand, thus tempting 
the \textit{Refiner} model to extract.
% This also helps explain \textit{Refiner}'s sub-optimal performance in ARC-Challenge task in \autoref{tab:refiner-vs-rag}.

% \input{limitations}

\section{Conclusion}
This work presents a novel paradigm for Retrieval-Augmented Generation (RAG), introducing \textit{Refiner}, a novel end-to-end paradigm that extracts and restructures document chunks, organizing query-relevant context-completed content into sections, thereby enhancing the performance of downstream language models. Thorough evaluations on five open-domain question-answering tasks conclusively demonstrate that \textit{Refiner} outperforms state-of-the-art RAG solutions that employ more complex enhancements, rivaling and surpassing concurrent prompt compress approaches with a significantly higher token compress rate. Our \textit{Refiner} has been tested in collaboration with various downstream language models and retrieval settings, proving to be resilient to the challenges posed by lengthy and noisy document chunks and consistently effective. Finally, \textit{Refiner} is a plug-and-play solution that can be seamlessly integrated with modular RAG systems, facilitating its application across diverse open-source frameworks.

\section{Limitations} \label{sec:Limitations}

This study seeks to enhance the quality of Large Language Model (LLM) outputs, a deficiency that continues to have significant real-world consequences, including spread of misinformation and provision of incorrect and dangerous advice. Notwithstanding our method and evaluation have demonstrated efficacy, they are conducted in a relatively indirect way, as how to evaluate the correctness of generated structural output from the model still requires further study. Moreover, our model is trained to extract and section information from open domain document chunks, its robustness has not been tested on alternative input structures, such as table data or domain-specific documents, which raises our concerns about potential misuse. Additionally, due to the nature of \textit{Refiner} being a probabilistic model, there is always a probability of extracting content that does not exactly derive from the document chunks. Therefore, when using them, it is still necessary to exercise caution when distinguishing and verifying the output.

\section{Broader Impact}
\label{sec:broader-impact}

Current RAG system has been found to hinder LLMs from producing misleading ~\cite{kwon2023efficient} or incorrect information~\cite{yan2024corrective}. Although there exist concurrent studies that adopt extractive and summarize models to distill information from retrieved documents in post-retrieval process, none have considered the relatedness of document-wise information, potentially resulting in a loss of contextual information that sometimes critical for downstream LMs to arrive at a correct answer. We believe our \textit{Refiner} is an important step to fill this gap. Furthermore, in contrast to the pre-retrieval and retrieval processes, the post-retrieval process has not been fully explored~\citep{gao2024retrievalaugmented} by academia and, in our opinion, holds significant potential. Therefore, we hope that our work will shed light on this process and inspire further research in this area. The \textit{Refiner} dataset, code and model files will be released under the CC BY-NC 4.0 license.

\section{Acknowledgement}
We thank Kunpeng Zhang for fruitful discussions in the early stages of this work. We thank Yibo Yan for valuable feedback on the paper, and hanying Sun, Fan Shi, jinghuai Ou for moral supports.

\clearpage
\newpage
{\small
\bibliography{references}
}

%%%%%%%%%%%%%%%%%%%%%%%%%%%%%%%%%%%%%%%%%%%%%%%%%%%%%%%%%%%%

\newpage

\onecolumn
\appendix
\addcontentsline{toc}{section}{Appendix} % Add the appendix text to the document TOC
\part{Appendix} % Start the appendix part

\parttoc % Insert the appendix TOC
\newpage

\section{Model Cards}
Below illustrates models inclusive in our work, all usages of models are under the compliance of their intended use.

\begin{table}[th!]
\centering
\small
\renewcommand\tabcolsep{1.0pt} % column space
\begin{tabular}{cp{12cm}}
    \toprule
    \textbf{Model Name} & \textbf{Description} \\
    \midrule
    \multirow{3}{*}{\parbox{2cm}{\centering GPT-3.5-Turbo}}
    & Developed by OpenAI, GPT-3.5-Turbo, or ChatGPT, is an advanced NLP model in the GPT series, featuring enhanced context understanding and text generation capabilities. Trained on a vast array of internet text data, it excels in tasks such as text generation, question answering, translation, and programming assistance, finding use in sectors like customer service, content creation, and education (URL: \url{https://openai.com/chatgpt}). \\
    \midrule
    \multirow{3}{*}{\parbox{1cm}{Meta-Llama3-Instruct \\ \centering(\text{8B, 70B})}} & 
    Meta-Llama3-Instruct, a cutting-edge language model designed to excel in understanding and responding to natural language instructions. This innovative AI model is capable of comprehending complex directives and generating human-like responses that are both accurate and informative. By leveraging advanced machine learning algorithms and large-scale training datasets, Meta-Llama3-Instruct has achieved unprecedented levels of language understanding and generation capabilities, making it an invaluable tool for a wide range of applications. \\
    % \midrule
    % \multirow{3}{*}{\parbox{2cm}{Llama2-Base \\ \centering(\text{7, 13, 70B})}}
    % & Llama2-Base~\citep{touvron2023llama}, a foundational model in Meta's Llama 2 series, represents a significant segment of LLMs. In contrast to Llama2-Chat, which is tailored for conversational applications, Llama2-Base embodies the broader foundational aspects of the series. The model's scale varies from 7B to 70B parameters, positioning it within the realm of pre-trained and fine-tuned text generation models. Llama2-Base is a versatile foundation, catering to various applications beyond just conversational contexts. With its extensive knowledge base and comprehension skills, it is adept at handling various text-related tasks like generation, summarization, translation, and question-answering. \\
    \midrule
    \multirow{3}{*}{\parbox{2cm}{Llama2-Chat \\ \centering(\text{7, 13, 70B})}}
    & Llama2-Chat~\citep{touvron2023llama}, developed and publicly released by Meta, is a refined version of LLMs, optimized specifically for conversational purposes. The Llama2-Chat belongs to the Llama 2 family, a series of pre-trained and fine-tuned generative text models with 7 to 70 billion parameters. Being a fine-tuned LLM, Llama2-Chat excels in dialogue scenarios. \\
    \midrule
    \multirow{3}{*}{\parbox{1cm}{Alpaca \\ \centering(\text{7B})}}
    & Alpaca~\citep{taori2023stanford} is a language model that has been fine-tuned from Meta's LLaMA 7B model for instruction-following tasks. Using 52,000 self-instruct style demonstrations, it was trained with OpenAI's text-davinci-003 model for instruction-following tasks. In evaluations, Alpaca has shown behaviors akin to OpenAI's text-davinci-003. Notably, Alpaca is distinguished by its small size, ease of replication, and low cost, making it an efficient, accessible model. \\
    \midrule
    \multirow{3}{*}{\parbox{2cm}{\centering{RECOMP\\Abstractive\\Compressor}}}
    & RECOMP~\citep{xu2023recomp} Abstractive Compressor is a part of the RECOMP (Retrieve, Compress, Prepend) method, which involving retrieving documents and compressing them into textual summaries before they are integrated into the context for the language model. This compression not only reduces computational costs but also makes it easier for the language model to identify relevant information in long retrieved documents. The RECOMP Abstractive Compressor is one of the two types of compressors used in this method, which generates summaries by synthesizing information from multiple documents. This allows the compressor to create a concise summary that contains the key information from the retrieved documents. The Abstractive Compressor is trained to improve the performance of the language model on end tasks when the generated summaries are prepended to the language model’s input, while keeping the summary concise. If the retrieved documents are irrelevant to the input or offer no additional information to the language model, the compressor can return an empty string. \\
    \midrule
    \multirow{3}{*}{\parbox{2.4cm}{LongLLMLingua}} & 
    LongLLMLingua~\citep{jiang-etal-2023-longllmlingua} is a method designed to compress prompts, thereby enhances the performance of LLMs, reduce computational and financial costs, and decrease latency. By compressing the prompt on token level, it reduces costs and boosts efficiency, improving performance by up to 21.4\% using only 1/4 of the tokens. Built upon LLMLingua~\citep{jiang-etal-2023-llmlingua}, the contributions of this paradigm are: (1) a question-aware coarse-to-fine compression method to enhance key information density in the prompt; (2) a document reordering mechanism to minimize information loss; (3) dynamic compression ratios to integrate coarse-grained and fine-grained compression for adaptive granular control; (4) a post-compression subsequence recovery strategy to improve key information integrity. \\
    \midrule
    \multirow{3}{*}{\parbox{2cm}{\centering{Self-RAG}}}
    & Self-Reflective Retrieval-Augmented Generation, or Self-RAG~\citep{asai2023selfrag}, is a framework designed to enhance the capabilities of LLMs. It does this by integrating retrieval and self-critique mechanisms into the model’s generation process. The framework trains a single arbitrary language model to adaptively retrieve passages on-demand, generate text, and reflect on retrieved passages and its own generations using special tokens, called reflection tokens. Reflection tokens are categorized into retrieval and critique tokens. Retrieval tokens indicate the need for retrieval, while critique tokens assess the quality of the generation1. Generating reflection tokens makes the language model controllable during the inference phase, enabling it to tailor its behavior to diverse task requirements. \\
    \midrule
    \multirow{3}{*}{\parbox{2cm}{\centering{RQ-RAG}}}
    & Following Self-RAG, RQ-RAG, or Learning to Refine Queries for Retrieval Augmented Generation~\citep{chan2024rqrag}, is a paradigm to refine queries with capabilities for explicit rewriting, decomposition, and disambiguation. By regenerating responses based on search results from external search engine, RQ-RAG surpasses the previous state-of-the-art by an average of 1.9\% across three single-hop QA datasets, and also demonstrates enhanced performance in handling complex, multi-hop QA datasets.
\\
    \bottomrule
\end{tabular}
\caption{Model Cards.}
\label{tab:model_cards}
\end{table}

\newpage
\section{Training Details}
\label{sec:training-details}

\subsection{Dataset}
\paragraph{Single-hop QA Dataset}consists of ARC Challenge~\citep{Clark2018ThinkYH} train, a set of natural, grade-school science questions, PubHealth~\citep{kotonya-toni-2020-explainable} train, a fact-checking for claims provided contents that support or refute to the claims, and TriviaQA~\citep{joshi2017triviaqa} train, a challenging reading comprehension QA dataset. Since ARC Challenge train and TriviaQA train dataset do not contain document candidates, we use off-the-shelf Contriever-MS MARCO to retrieve top document chunks from 2020 English Wikipedia using official Wikipedia embeddings.

\paragraph{Multi-hop QA Dataset}includes HotpotQA~\citep{yang2018hotpotqa} train dataset. The dataset contains 90447 pairs of QA and a set of supporting contents randomly mingled with irrelevant or misleading contents.

We further clean datasets by removing answers that are absent in the contents or in the outputs of teacher models. Please refer to \autoref{tab:train_dataset} for dataset statistics.

\subsection{Teacher Models}
We employ five teacher models to generate the proposed structured output for each query-document pair, including Llama2-70B-Chat~\citep{touvron2023llama}, Meta-Llama3-(8B,70B)-Instruct and Alpaca-7B~\citep{taori2023stanford}. We design four prompts to tackle with diverse QA tasks.
For each output generated by each teacher model, we utilize regular expressions to parse and resolve sections, titles, and contents. We then exclude any contents that are not present in the provided document chunks. The remaining contents are subject to a voting process among the teacher models, whereby only those contents that receive a majority vote from the teacher models are retained. Finally, the filtered contents are concatenated to formulate the final training dataset. The detailed task-specific prompts are illustrated in \autoref{tab:teacher_model_prompts}, and hyperparameters of the teacher models are listed in \autoref{tab:teacher-models}.

\begin{table}[htbp]
    \centering
    % \resizebox{\textwidth}{!}
    % {
    \begin{tabular}{ccccc}
    \toprule
        \textbf{Hyperparameters}  & \textbf{Meta-Llama3-70B-Instruct} & \textbf{Meta-Llama3-8B-Instruct} \\ \hline
        temperature & None & None \\ 
        repetition-penalty & 1.0 & 1.0 \\ 
        max-length & 2048 & 2048 \\ 
        do-sample & False & False \\ 
    \midrule
        \textbf{Hyperparameters} & \textbf{Llama2-70B-Chat} & \textbf{Llama2-13B-Chat} & \textbf{Alpaca-7B} \\ \hline
        temperature & None & None & None  \\ 
        repetition-penalty & 1.0 & 1.0 & 1.0  \\ 
        max-length & 2048 & 2048 & 2048 \\ 
        do-sample & False & False & False \\ 
    \bottomrule
    \end{tabular}
    % }
    \caption{Hyper-parameters of teacher models.}
\label{tab:teacher-models}
\end{table}

\subsection{Structure Correction}

Once outputs are generated by teacher models, we make a series of cleaning and filtering procedures so as to satisfy the proposed principles. Firstly, we filter out contents that are absent in retrieved document chunks, store legitimate sections voted by teacher models, then remove sentences that are vetoed by teacher models. Finally, the selected sections are unified by re-aligning section numbers, titles and extractive contents to form the \textit{Refiner}’s training data as illustrated in \autoref{alg:train-process}.

\subsection{Environment and Hyperparameters}
\label{train-env-hyperparam}
We use 32GB memory and 4 Nvidia A100 with 80GB memory to train our model. 
We adopt distributed data parallel~\citep{li2020pytorch} and Deepspeed stage 3~\citep{10.5555/3433701.3433727} for multi-GPU distributed training. We apply parameter-efficient funetuning with LoRA~\citep{hu2021lora} rank of 64, alpha of 16 on Llama2-7B-Chat. AdamW~\citep{loshchilov2019decoupled} is adopted as optimizer. All seeds are set to 633. The total train steps reaches 6 hours in the specified environment.

\subsection{Training Process}
\label{sec:train-details}
As shown in \autoref{tab:train_dataset}, the cleaned datasets are highly imbalanced in number of samples. Thus, in order to train the model effectively, we practice progressive training: First, we form training dataset with TriviaQA and HotpotQA dataset combined, the model is trained for 3 epochs with a batch size of 128, a peak learning rate of 4e-5 with 3\% warmup steps, and linear decay afterward.
Subsequently, we continue training the model combining Arc Challenge and PubHealth datasets, with 5 epoch and a batch size of 128, a peak learning rate of 2e-5 with 5\% warmup steps.

\begin{figure}

\begin{framed}
\small
\textbf{Input:} Given $n$ Teacher LM $\{M_j\}_1^n$, \textit{Refiner} $\mathcal{M_{\theta}}$, query set $\{q_i\}_1^N$, where ${q}$ represents the user's query, $R$ denotes retrieval system by default, $C$ denotes corpus set, $\mathcal{D}$ is a set of corresponding retrieved documents such that $\mathcal{D} \in C$, $\mathbf{V}$ is a collection of parsed contents in sentence level generated by teacher LM. $\mathcal{T}$ denotes temporary content collection.\\
\textbf{Output:} An updated \textit{Refiner} $\mathcal{M_{\theta}}$ \\
% \vspace{-0.6em}
\begin{algorithmic}[1]
\STATE $\mathcal{S} \gets \emptyset$
\FOR {$i  \in \{1, \dots, N \}$} 
    \STATE {$\mathbf{D}_i = \textbf{TopN}(R(q_i,C))$}
    \FOR {$j  \in \{1, \dots, n \}$} \label{alg:start_it}
        \STATE{ $\mathbf{s_j} = \textbf{Generate}(M_j, [{q_i},\mathbf{D}_{i}])$ }\label{line:decode}
        \STATE{ $\mathbf{V} = \textbf{ParseGeneratedContent}(\mathbf{s_j})$ }
        % \IF {$\mathbf{V} = \emptyset$}
        %     \STATE{ $\textbf{continue}$ }
        % \ENDIF

        \FOR {$\textbf{content}  \in \mathbf{V}$} 
            \IF {$\textbf{content} \not\in \mathbf{D}_{i} $ }
                % \STATE{${v} = \textbf{restructure}(\mathbf{v})$ }
                \STATE{$\mathbf{V.remove}(\textbf{content}) $}
            \ENDIF
        \ENDFOR

    \ENDFOR

    \STATE {$\mathcal{T} \gets \emptyset$}
    \STATE{$c = \textbf{Counter}(\mathbf{V})$}

        \FOR{$\textbf{content}, \textbf{count} \in c$}
        \STATE{// Iterate over result from counter to get \\ // number of occurrence for each content}
            \IF {$ \textbf{count} > n/2$ }
                \STATE{${\textbf{content}} = \textbf{assignSection}(\textbf{content})$}
                \STATE{$\mathcal{T} \gets \mathcal{T} \cup \textbf{content} $}
            \ENDIF
        \ENDFOR

    \STATE{$\mathcal{Y}_i \gets \textbf{Concatenate}(\mathcal{T})$}
    \STATE{$S \gets S \cup [{q_i},\mathbf{D}_i; \mathcal{Y}_i] $}

\ENDFOR

\STATE {$\mathcal{M_{\theta}}=\textbf{Finetune}(\mathcal{M_{\theta}}, S)$}

\end{algorithmic}
\end{framed}

\caption{Prepare curated train dataset for \textit{Refiner}.}\label{alg:train-process}

\end{figure}

\begin{table}[htbp]
    \centering
    \begin{tabular}{ccc}
    \toprule
        Dataset & Initial & Cleaned \\
    \midrule
        Arc Challenge & 1119 & 1119 \\
        PubHealth & 9513 & 9513 \\
        TriviaQA & 61888 & 56655 \\
        HotpotQA & 90447 & 79157 \\
    \bottomrule
    \end{tabular}
    \caption{Training data statistics.}
    \label{tab:train_dataset}
\end{table}

\begin{table*}[ht!]
\begin{tabular}{p{1.5cm}p{12cm}}
\toprule
\textbf{Dataset} & \textbf{Prompts} \\ 
\midrule
TriviaQA \&\newline HotpotQA & \textbf{\textit{System}}: You are an expert research assistant. Your job is to find the quotes from the markdown documents that are relevant to a question.\texttt{\textbackslash n} Please mark quotes with sections and titles of documents to group quotes by different information only from relevant documents.\texttt{\textbackslash n} Rule of labelling sections: if the fact of the first dummy quote "aaa" from a document with title "\#\# AAA" is consistent with that of the second dummy quote "bbb" from a document with title "\#\# BBB", and they contradict to that of the third dummy quote "ccc" from a document with title "\#\# CCC", then label them as:\texttt{\textbackslash n} ```\texttt{\textbackslash n}1.1. \#\# AAA\texttt{\textbackslash n} aaa\texttt{\textbackslash n}\texttt{\textbackslash n}1.2. \#\# BBB\texttt{\textbackslash n} bbb\texttt{\textbackslash n}\texttt{\textbackslash n}2.1. \#\# CCC\texttt{\textbackslash n} ccc\texttt{\textbackslash n}\texttt{\textbackslash n}```\texttt{\textbackslash n} Quotes should be verbatim and context completed.\texttt{\textbackslash n} Please respond without any explanation. \newline\newline \textbf{\textit{User}}: Question: \{\texttt{query}\}\texttt{\textbackslash n} Document (multiple documents are separated by "---"):\texttt{\textbackslash n} \{\texttt{context}\} \\
\midrule
ARC Challenge & \textbf{\textit{System}}: You are an expert research assistant. Your job is to find the quotes from the markdown documents that are relevant to a question.\texttt{\textbackslash n} Please mark quotes with sections and titles of documents to group quotes by different information only from relevant documents that either support or contradict to the question.\texttt{\textbackslash n} Rule of labelling sections: if the fact of the first dummy quote "aaa" from a document with title "\#\# AAA" is consistent with that of the second dummy quote "bbb" from a document with title "\#\# BBB", and they contradict to that of the third dummy quote "ccc" from a document with title "\#\# CCC", then label them as:\texttt{\textbackslash n} ```\texttt{\textbackslash n}1.1. \#\# AAA\texttt{\textbackslash n} aaa\texttt{\textbackslash n}\texttt{\textbackslash n}1.2. \#\# BBB\texttt{\textbackslash n} bbb\texttt{\textbackslash n}\texttt{\textbackslash n}2.1. \#\# CCC\texttt{\textbackslash n} ccc\texttt{\textbackslash n}\texttt{\textbackslash n}```\texttt{\textbackslash n} Quotes should be verbatim and context completed.\texttt{\textbackslash n} Please respond without any explanation.\newline\newline \textbf{\textit{User}}: Question: \{\texttt{query}\}\texttt{\textbackslash n} Document (multiple documents are separated by "---"):\texttt{\textbackslash n} \{\texttt{context}\} \\ 
\midrule
PubHealth & \textbf{\textit{System}}: You are an expert research assistant. Your job is to find the quotes from the markdown documents that either support or contradict to a statement.\texttt{\textbackslash n} Please mark quotes with sections and titles of documents to group quotes by different information only from relevant documents that are helpful with answering to the statement.\texttt{\textbackslash n} Rule of labelling sections: if the fact of the first dummy quote "aaa" from a document with title "\#\# AAA" is consistent with that of the second dummy quote "bbb" from a document with title "\#\# BBB", and they contradict to that of the third dummy quote "ccc" from a document with title "\#\# CCC", then label them as:\texttt{\textbackslash n} ```\texttt{\textbackslash n}1.1. \#\# AAA\texttt{\textbackslash n} aaa\texttt{\textbackslash n}\texttt{\textbackslash n}1.2. \#\# BBB\texttt{\textbackslash n} bbb\texttt{\textbackslash n}\texttt{\textbackslash n}2.1. \#\# CCC\texttt{\textbackslash n} ccc\texttt{\textbackslash n}\texttt{\textbackslash n}```\texttt{\textbackslash n} Quotes should be verbatim and context completed.\texttt{\textbackslash n} Please respond without any explanation.\newline\newline \textbf{\textit{User}}: Statement: \{\texttt{query}\}\texttt{\textbackslash n} Document (multiple documents are separated by "---"):\texttt{\textbackslash n} \{\texttt{context}\}  \\
\bottomrule
\end{tabular} \vspace{-0.3em}
\caption{Prompts used to generated structured outputs from teacher models \{\texttt{query}\} and \{\texttt{context}\} represent the actual input query and retrieved documents.}
\label{tab:teacher_model_prompts}
\end{table*}

\clearpage
\newpage
\section{Evaluation Details}
\label{sec:evaluation-details}

\subsection{Hyperparameters for Adavanced RAG}
\label{sec:hyperparams-for-advanced-rag}
For Self-RAG, we adopt the default inference settings. Specifically, we set the weight terms $\textbf{IsRel}$, $\textbf{IsSup}$, $\textbf{IsUse}$ values of 1.0, 1.0 and 0.5, respectively. The retrieval threshold is set to 0.2 to encourage frequent retrieval.

\subsection{Hyperparameters for Concurrent Compressor}
\label{sec:hyperparams-for-concurrent-compressor}
We assess the performance of concurrent compressors encompassing RECOMP abstractive compressor~\citep{xu2023recomp} and LongLLMLingua~\citep{jiang-etal-2023-longllmlingua}.
Notably, the RECOMP abstractive compressor has two versions, one trained on TriviaQA dataset and the other on HotpotQA dataset. We evaluate the former version on the PopQA, TriviaQA, and ARC Challenge datasets, while the latter version was tested on the HotpotQA and 2WikiMultihop datasets.
Regarding the hyperparameters of LongLLMLingua, we utilized a default compression rate of 0.5 and a dynamic context compression ratio of 0.3.

\subsection{Environment and Hyperparameters}
\label{evaluate-env-hyperparam}
We use 32GB memory and 4 Nvidia A100 with 80GB memory for inference. we adopt vllm~\citep{kwon2023efficient} to speed up inference. All models, including downstream LMs and \textit{Refiner}, use greedy algorithm to sample their next tokens, and max new tokens is set to 2048.

\subsection{Prompt Template for \textit{Refiner}}
To efficiently train \textit{Refiner} and accelerate convergence speed, we revise its base model, Llama2-Chat's chat template, which is illustrated in \autoref{tab:refiner_prompt}.

\begin{table*}[ht!]

\begin{tabular}{p{3cm}p{10cm}}
\toprule
\textbf{Prompt Template for \textit{Refiner}} & \text{[INST]<<SYS>>[MONITOR]\{\texttt{documents}\}<</SYS>>\{\texttt{query}\}[/INST]} \\
\midrule
\textbf{Prompt Template for {\{\texttt{documents}\}}} & \#\# \{\texttt{title}\}\newline \{\texttt{content}\}\newline
\#\# \{\texttt{title}\}\newline \{\texttt{content}\}\newline
\#\# \{\texttt{title}\}\newline \{\texttt{content}\}\newline 
...\\
\bottomrule
\end{tabular} \vspace{-0.3em}
\caption{Prompt template for \textit{refiner} where \{\texttt{query}\} represents the actual user input query, \{\texttt{documents}\} represents document chunks retrieved for extraction, \{\texttt{content}\} represents content in document and \{\texttt{title}\} represents title of the document from which the content derives.}
\label{tab:refiner_prompt}
\end{table*}

\subsection{Prompt Template for Downstream LM}
To simulate real-world application of \textit{Refiner}, we prompt downstream LMs to work with \textit{Refiner}'s output as illustrated in the \autoref{tab:evaluate_prompt}.

\begin{table*}[ht!]
\centering
\begin{tabular}{p{13cm}}
\toprule
\textbf{Prompt} \\ 
\midrule
\textbf{\textit{System}}: You are an AI assistant backboned by selective content from different documents, answer user's question helpfully and precisely, with the guidance of the following steps:\newline * If there are no content provided: determine whether it is still possible to answer precisely to the question.\newline * If is possible, offer a helpful answer. Otherwise, offer the reason of impossibility.\newline \newline * If there exists contents: determine whether the necessary information to answer the question is either directly mentioned or can be inferred from the documents.\newline * When there exists different information that can answer to the question, determine whether it is the question too opaque that causes the problem.\newline * If not, answer with a summarized information. Otherwise, also provide advice or ask question to disambiguate.\newline * When summarizing, ensure to include contents that are relevant to the question.\newline Here is the content:\newline
 \{\texttt{refiner}\} \newline
 \textbf{\textit{User}}: \{\texttt{query}\} \\
\bottomrule
\end{tabular} \vspace{-0.3em}
\caption{Prompts used to generated structured outputs from teacher models \{\texttt{query}\} represent the actual input query, and \{\texttt{refiner}\} represent the actual generated output from \textit{Refiner}.}
\label{tab:evaluate_prompt}
\end{table*}

\clearpage
\newpage
\subsection{Evaluation Metric for ARC Challenge}
\label{acc-for-arc_c}
\subsubsection{Motivation}
To deliver our motivation to revising evaluation metric for ARC Challenge task, we provide a case from the ARC Challenge test subset.In this case, the model is prompted to answer question based on the A, B, C and D choices given, while through reading the context we can easily find out that the model is answering with the choice B. However, the generated content contains all of the four choices, as previous work does not take such cases into account, the performance of the model could be over-evaluated.

\begin{table}[!htb]
\begin{tabular}{@{}p{1.2cm}|p{6.5cm}|p{3.4cm}|p{1.8cm}}

\toprule
% \textbf{Dataset} & \textbf{Prompts} \\ 
\textbf{Method} & \textbf{Case in Arc Challenge test subset} & \textbf{Result}&\textbf{Comment}\\
\midrule
\multirow{4}{1.2cm}{Previous Work}  & \multirow{8}{6.5cm}{\textbf{Question}: The morning temperature in a city is 41°F. If a sunny, mild day is forecast, which temperature is most likely for 2:00 p.m.? \newline A. 32° F\newline B. 41° F\newline C. 78° F\newline D. 98° F \newline \textbf{Ground Truth}:  \underline{C} \newline\newline \textbf{LLM outputs}: Based on the given information, the best answer choice is ({\color{red-color}B}) 41°F.\newline\newline The given information states that the morning temperature is 41°F, which suggests that the day will be sunny and mild. As the temperature drops gradually throughout the day, it is unlikely that the temperature at 2:00 p.m. will be above 78°F or 98°F, as these temperatures are much higher than the morning temperature. Therefore, the most likely temperature for 2:00 p.m. is 41°F, which is the same as the morning temperature.\newline\newline Answer choice ({\color{red-color}A}) 32°F is too low, as it is below the morning temperature and unlikely to be the temperature at 2:00 p.m. in a sunny and mild day. Answer choice ({\color{red-color}C}) 78°F is also unlikely, as it is higher than the morning temperature and may not be reached in the afternoon. Answer choice ({\color{red-color}D}) 98°F is the highest temperature mentioned in the passage and is therefore unlikely to be the temperature at 2:00 p.m.} & \multirow{4}{3.4cm}{\textbf{Matches ground truth C in generated sentence:} "Answer choice ({\color{red-color}C}) 78°F is also unlikely"} & \multirow{4}{1.8cm}{Wrong evaluation result} \\
& & & \\
& & & \\
& & & \\
& & & \\
& & & \\
& & & \\
& & & \\
& & & \\
& & & \\
& & & \\
& & & \\
& & & \\
& & & \\
& & & \\
& & & \\
& & & \\
& & & \\
\cline{1-1} \cline{3-4}\\
\multirow{4}{1.2cm}{Ours} & \multirow{8}{6.5cm}{} & \multirow{4}{3.4cm}{\textbf{Ground truth C mismatches with predicted B in sentence}: "Based on the given information, the best answer choice is ({\color{red-color}B}) 41°F."} & \multirow{4}{1.8cm}{Correct evaluation result} \\
& & & \\
& & & \\
& & & \\
& & & \\
& & & \\
& & & \\
& & & \\
& & & \\
& & & \\
& & & \\
& & & \\
& & & \\
& & & \\
& & & \\
& & & \\
& & & \\
\bottomrule

\end{tabular}
\caption{The result of regular-expression-based accuracy compared to previous work.}
\end{table}

\subsubsection{Code Comparison}
For a concise convey of our evaluation metric, we disclose previous work and our pseudo-python code in the following list.

\begin{lstlisting}[escapechar=!]
# Previous Methods
def match(prediction: str, ground_truth: list[str]):
    # Argument 'prediction' represents generated output from downstream LM,
    # and 'ground_truth' represents a list of answer key strings.
    # For example, if the ground truth choice for a question is A,
    # then ['A'] will be passed to ground_truth.

    for gt in ground truth:
        if gt in prediction:
            return 1

    return 0
\end{lstlisting}
\begin{lstlisting}[escapechar=!]
# Our Methods
import re


def arc_c_acc(prediction: str, ground_truth: list[str]):
    # The presence of 'is' preceding the letter 'A' to 'E'
    # is a necessary condition for a match to occur.
    regex_answer_key = re.compile(r"is[^\w]+([A-E])[^\w]+", re.DOTALL)

    # re.search only looks for the first match occurence.
    if len(ground_truth) == 0:
        return re.search(regex_answer_key, prediction) is None

    lst_res = []
    for key in ground_truth:
        search = re.search(regex_answer_key, prediction)
        if search is None:
            lst_res.append(False)
            continue

        if key.lower() == search.groups()[0].lower():
            lst_res.append(True)
        else:
            lst_res.append(False)

    # only when all ground truths are matched should we return True.
    return all(lst_res)
\end{lstlisting}

% \begin{tabular}{cccc} 
% \hline
% \textbf{Method} & \textbf{Case} & \textbf{Metric Function }&\textbf{value}\\ 
% \hline
% \multirow{2}{2cm}{self-rag} & \multirow{4}{8cm}{ Which of the following statements about Barack Obama are correct?A. Barack Obama was the first African American president of the United States.B. Barack Obama earned a Doctor of Law degree from Harvard University.C. Barack Obama won the presidential elections in 2008 and 2016.D. Barack Obama was born in Hawaii and spent part of his childhood in Indonesia.Answer: A and D} & \multirow{2}{2cm}{match}&\multirow{2}{2cm}{right} \\
% \multirow{2}{2cm}{our} & & \multirow{2}{2cm}{calc\_acc }& \multirow{2}{2cm}{wrong}\\
% \bottomrule
% \end{tabular}

\clearpage
\newpage
\section{Ablation Analysis Details}
\label{sec:ablation-detail}
We present a comprehensive disclosure of all ablation analysis results, with specific sections omitted from the \textit{Refiner}'s output. The accuracy outcomes under the top 5 retrieval setting are detailed in \autoref{tab:ablation_in_model_top5}, while the accuracy outcomes under the top 10 retrieval setting are presented in \autoref{tab:ablation_in_model_top10}.

\begin{table*}[!htb]
    \centering
    \begin{tabular}{lcc|c}
    \toprule
        \textbf{Task} & \textit{Refiner} & acc. w/o. Section & acc. Decay\\
    \midrule
    \multicolumn{3}{l}{\textit{\textbf{Llama2-7B-Chat}}} \\
        PopQA & 59.3 & 57.7 & 1.6 \\
        TriviaQA & 68.9 & 66.5 & 2.4 \\
        ARC Challenge & 51.1 & 51.7 & -0.6 \\
        HotpotQA & 47.4 & 47.3 & 0.1 \\
        2WikiMultihop & 46.0 & 46.3 & -0.2 \\
    \midrule
    \multicolumn{3}{l}{\textit{\textbf{Llama2-13B-Chat}}} \\
        PopQA & 59.0 & 58.5 & 0.5 \\
        TriviaQA & 70.5 & 69.4 & 1.1 \\
        ARC Challenge & 54.2 & 54.4 & -0.2 \\
        HotpotQA & 49.7 & 49.5 & 0.2 \\
        2WikiMultihop & 49.7 & 47.0 & -1.3 \\
    \midrule
    \multicolumn{3}{l}{\textit{\textbf{Llama2-70B-Chat}}} \\
        PopQA & 58.3 & 58.0 & 0.3\\
        TriviaQA & 72.2 & 71.4 & 0.8 \\
        ARC Challenge & 55.9 & 58.7 & -2.8 \\
        HotpotQA & 52.3 & 52.2 & 0.1 \\
        2WikiMultihop & 50.1 & 50.2 & -0.1 \\
    \midrule
    \multicolumn{3}{l}{\textit{\textbf{Meta-Llama3-8B-Instruct}}} \\
        PopQA & 58.3 & 57.9 & 0.4 \\
        TriviaQA & 69.1 & 66.8 & 2.3 \\
        ARC Challenge & 71.8 & 74.4 & -2.6 \\
        HotpotQA & 49.0 & 49.2 & -0.2 \\
        2WikiMultihop & 48.0 & 48.0 & 0.0 \\
    \midrule
    \multicolumn{3}{l}{\textit{\textbf{Meta-Llama3-70B-Instruct}}} \\
        PopQA & 57.9 & 57.6 & 0.3 \\
        TriviaQA & 70.9 & 68.9 & 2.0 \\
        ARC Challenge & 84.1 & 85.5 & -1.4 \\
        HotpotQA & 49.2 & 48.7 & 0.5 \\
        2WikiMultihop & 46.7 & 46.8 & -0.1 \\
    \bottomrule
    \end{tabular}
    \caption{\textit{Refiner}-augmented model accuracy under top 5 retrieval setting.}
    \label{tab:ablation_in_model_top5}
\end{table*}

\begin{table*}[!htb]
    \centering
    \begin{tabular}{lcc|c}
    \toprule
        \textbf{Task} & \textit{Refiner} & acc. w/o. Section & acc. Decay\\
    \midrule
    \multicolumn{3}{l}{\textit{\textbf{Llama2-7B-Chat}}} \\
        PopQA & 59.3 & 58.4 & 0.9 \\
        TriviaQA & 68.9 & 66.2 & 2.7 \\
        ARC Challenge & 51.8 & 50.4 & 1.4 \\
        HotpotQA & 65.1 & 63.2 & 1.9 \\
        2WikiMultihop & 61.0 & 57.6 & 3.4 \\
    \midrule
    \multicolumn{3}{l}{\textit{\textbf{Llama2-13B-Chat}}} \\
        PopQA & 59.4 & 57.8 & 1.5 \\
        TriviaQA & 70.6 & 68.3 & 2.3 \\
        ARC Challenge & 55.1 & 55.0 & 0.1 \\
        HotpotQA & 67.1 & 66.3 & 0.8 \\
        2WikiMultihop & 61.6 & 61.4 & 0.2 \\
    \midrule
    \multicolumn{3}{l}{\textit{\textbf{Llama2-70B-Chat}}} \\
        PopQA & 60.8 & 58.2 & 2.6 \\
        TriviaQA & 72.1 & 70.9 & 1.2 \\
        ARC Challenge & 57.0 & 59.8 & -2.8 \\
        HotpotQA & 69.8 & 69.0 & 0.8 \\
        2WikiMultihop & 68.3 & 68.1 & 0.2 \\
    \midrule
    \multicolumn{3}{l}{\textit{\textbf{Meta-Llama3-8B-Instruct}}} \\
        PopQA & 59.7 & 57.8 & 1.9 \\
        TriviaQA & 69.3 & 67.2 & 2.1 \\
        ARC Challenge & 71.1 & 73.3 & -2.2 \\
        HotpotQA & 67.2 & 67.4 & -0.2 \\
        2WikiMultihop & 63.7 & 63.6 & 0.1 \\
    \midrule
    \multicolumn{3}{l}{\textit{\textbf{Meta-Llama3-70B-Instruct}}} \\
        PopQA & 59.8 & 58.5 & 1.3 \\
        TriviaQA & 71.1 & 69.4 & 1.7 \\
        ARC Challenge & 84.7 & 84.4 & 0.3 \\
        HotpotQA & 68.6 & 68.3 & 0.3 \\
        2WikiMultihop & 61.8 & 61.8 & 0.0 \\
    \bottomrule
    \end{tabular}
    \caption{\textit{Refiner}-augmented model accuracy under top 10 retrieval setting.}
    \label{tab:ablation_in_model_top10}
\end{table*}

\clearpage
\newpage
\section{More Analysis on Output Structure}
\label{sec:more-analysis-on-output-structure}
\begin{figure}[!htb]
    \centering
    \includegraphics[width=0.75\linewidth]{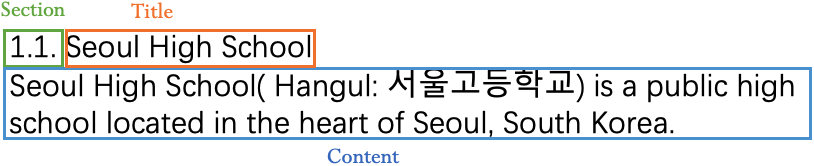}

    \caption{The structure of output generated by \textit{Refiner} consists of three parts: {\color{green-color}section}, {\color{orange-color}Title} and {\color{blue-color}content}, whereSection elucidates relatedness information among extracted contents, Title represents the Title of the Original document, content represents the question-relevant content extracted from that document chunk.}
    \label{fig:refiner-output-structure}
\end{figure}

Our \textit{Refiner} comes with a structured output, this allows for seamless dissemble of section, titles and contents (see \autoref{fig:refiner-output-structure}) using regular expressions, enables us to easily convert them into numerous alternative structures. In this context, we will explore a total of 16 additional different structure options, as illustrated in \autoref{tab:more-output-structures}.

\begin{table}[!htb]
\vspace{-0.6em}
\setlength{\tabcolsep}{4pt}
\begin{center}
\begin{tabular}{@{}cccl@{}}
\toprule

\textbf{Content Structure} & \textbf{Section Structure} & \multicolumn{1}{c}{\textbf{Title Structure}} & \multicolumn{1}{c}{\textbf{Template}} \\
 
\hline
& & \textbf{Original} & {\color{green-color}1.1.} ${\color{orange-color}[Title]}\texttt{\textbackslash n}{\color{blue-color}[Content]}$ \\
& \textbf{Hierarchy} & \textbf{Markdown} & {\color{green-color}1.1. }${\color{orange-color}\text{\#\# } [Title]}\texttt{\textbackslash n}{\color{blue-color}[Content]}$ \\
& & \textbf{Quote} & {\color{green-color}1.1.} ${\color{orange-color}"[Title]"}\texttt{\textbackslash n}{\color{blue-color}[Content]}$ \\

\cline{2-4}
& & \textbf{Original} & {\color{green-color}1.} ${\color{orange-color}[Title]}\texttt{\textbackslash n}{\color{blue-color}[Content]}$ \\
\makecell{\textbf{Original}} & \textbf{Numbered } & \textbf{Markdown} & {\color{green-color}1.}  ${\color{orange-color}\text{\#\# } [Title]}\texttt{\textbackslash n}{\color{blue-color}[Content]}$ \\
& & \textbf{Quote} & {\color{green-color}1.} ${\color{orange-color}"[Title]"}\texttt{\textbackslash n}{\color{blue-color}[Content]}$ \\

\cline{2-4}
& & \textbf{Original} & {\color{green-color}*} ${\color{orange-color}[Title]}\texttt{\textbackslash n}{\color{blue-color}[Content]}$ \\
& \textbf{Star} & \textbf{Markdown} & {\color{green-color}*} ${\color{orange-color}\text{\#\# }[Title]}\texttt{\textbackslash n}{\color{blue-color}[Content]}$ \\
& & \textbf{Quote} & {\color{green-color}*} ${\color{orange-color}"[Title]"}\texttt{\textbackslash n}{\color{blue-color}[Content]}$ \\

\hline
& & \textbf{Original} & {\color{green-color}1.1.} ${\color{orange-color}[Title]}\texttt{\textbackslash n}{\color{blue-color}"[Content]"}$ \\
& \textbf{Hierarchy} & \textbf{Markdown} & {\color{green-color}1.1.} ${\color{orange-color}\text{\#\# }[Title]}\texttt{\textbackslash n}{\color{blue-color}"[Content]"}$ \\
& & \textbf{Quote} & {\color{green-color}1.1.} ${\color{orange-color}"[Title]"}\texttt{\textbackslash n}{\color{blue-color}"[Content]"}$ \\

\cline{2-4}
& & \textbf{Original} & {\color{green-color}1.} ${\color{orange-color}[Title]}\texttt{\textbackslash n}{\color{blue-color}"[Content]"}$ \\
\textbf{Quote} & \textbf{Numbered } & \textbf{Markdown} & {\color{green-color}1.} ${\color{orange-color}\text{\#\# }[Title]}\texttt{\textbackslash n}{\color{blue-color}"[Content]"}$ \\
& & \textbf{Quote} & {\color{green-color}1.} $"{\color{orange-color}[Title]}"\texttt{\textbackslash n}{\color{blue-color}"[Content]"}$ \\

\cline{2-4}
& & \textbf{Original} & {\color{green-color}*} ${\color{orange-color}[Title]}\texttt{\textbackslash n}{\color{blue-color}"[Content]"}$ \\
& \textbf{Star} & \textbf{Markdown} & {\color{green-color}*} ${\color{orange-color}\text{\#\# }[Title]}\texttt{\textbackslash n}{\color{blue-color}"[Content]"}$ \\
& & \textbf{Quote} & {\color{green-color}*} ${\color{orange-color}"[Title]"}\texttt{\textbackslash n}{\color{blue-color}"[Content]"}$ \\

\bottomrule
\end{tabular} 
\end{center}\vspace{-0.6em}
\caption{Several restructure templates are utilized to assess the impact on the performance of downstream LMs. In these templates, $[Title]$ denotes the actual Title of the Original document, while $[Content]$ represents the actual content extracted from the corresponding document chunk. A comparative analysis is conducted between the Original hierarchicalSection structure and two alternative structures: a Numbered edSection structure and a Star Section structure. Furthermore, the Original Title is compared in its Original form and when formatted using markdown or quotation marks, while the Original content is compared in its Original form and when decorated with quotation marks.}
\label{tab:more-output-structures}
\vspace{-0.1in}
\end{table}

\begin{table}[!htb]
\vspace{-0.6em}
\setlength{\tabcolsep}{4pt}
\begin{center}
    \begin{tabular}{c|cc|cc|cc}
    \toprule
         \multirow{2}{*}{\textbf{Output Structure}} & \multicolumn{2}{c|}{\textbf{POPQA}} & \multicolumn{2}{c|}{\textbf{TQA}} & \multicolumn{2}{c}{\textbf{ARC-C}} \\ 
         \cmidrule(lr){2-3} \cmidrule(lr){4-5} \cmidrule(lr){6-7} 
         & mean & std & mean & std & mean & std \\
         \midrule
        \multicolumn{7}{c}{\textbf{Original Content}}  \\
        \midrule
        \multicolumn{1}{l|}{\textbf{Numbered Section Markdown Title}} & 1.0\% & 0.9\% & 1.7\% & 0.5\% & -0.8\% & 2.5\% \\
        \multicolumn{1}{l|}{\textbf{Numbered Section Original Title}} & 1.2\% & 0.8\% & 1.9\% & 0.5\% & -0.1\% & 1.4\% \\
        \multicolumn{1}{l|}{\textbf{Numbered Section Quote Title}} & 1.1\% & 0.8\% & 1.9\% & 0.6\% & -0.4\% & 2.2\% \\
        \multicolumn{1}{l|}{\textbf{Hierarchy Section Markdown Title}} & 1.1\% & 0.7\% & 1.7\% & 0.5\% & -1.2\% & 2.6\% \\
        \multicolumn{1}{l|}{\textbf{Hierarchy Section Quote Title}} & 1.0\% & 0.9\% & 1.8\% & 0.7\% & -0.5\% & 2.8\% \\
        \multicolumn{1}{l|}{\textbf{Star Section Markdown Title}} & 1.4\% & 0.7\% & 1.7\% & 0.6\% & -0.9\% & 2.5\% \\
        \multicolumn{1}{l|}{\textbf{Star Section Original Title}} & 1.2\% & 0.7\% & 1.9\% & 0.5\% & 0.1\% & 1.4\% \\
        \multicolumn{1}{l|}{\textbf{Star Section Quote Title}} & 0.9\% & 0.9\% & 1.9\% & 0.7\% & -0.5\% & 1.9\% \\
        \midrule
        \multicolumn{7}{c}{\textbf{Quote Content}}  \\
        \midrule
        \multicolumn{1}{l|}{\textbf{Numbered Section Markdown Title}} & 1.3\% & 0.7\% & 1.7\% & 0.6\% & -0.9\% & 2.6\% \\
        \multicolumn{1}{l|}{\textbf{Numbered Section Original Title}} & 1.2\% & 0.6\% & 1.8\% & 0.5\% & -0.3\% & 2.0\% \\
        \multicolumn{1}{l|}{\textbf{Numbered Section Quote Title}} & 1.4\% & 0.7\% & 1.8\% & 0.6\% & -0.9\% & 2.9\% \\
        \multicolumn{1}{l|}{\textbf{Hierarchy Section Markdown Title}} & 1.3\% & 0.7\% & 1.6\% & 0.7\% & -1.4\% & 3.2\% \\
        \multicolumn{1}{l|}{\textbf{Hierarchy Section Quote Title}} & 1.1\% & 0.7\% & 1.7\% & 0.7\% & -0.5\% & 2.9\% \\
        \multicolumn{1}{l|}{\textbf{Star Section Markdown Title}} & 1.2\% & 0.6\% & 1.7\% & 0.6\% & -1.2\% & 1.9\% \\
        \multicolumn{1}{l|}{\textbf{Star Section Original Title}} & 1.2\% & 0.6\% & 1.8\% & 0.6\% & -0.6\% & 1.7\% \\
        \multicolumn{1}{l|}{\textbf{Star Section Quote Title}} & 1.1\% & 0.7\% & 1.8\% & 0.7\% & -0.6\% & 2.4\% \\
    \bottomrule
    \end{tabular}
\end{center}\vspace{-0.6em}
\caption{Statistics of improved accuracy margin from various downstream LMs augmented by restructured outputs from \textit{Refiner}, compared with no structured output in single-hop tasks.}
\label{tab:more-output-structure-single-hop-result}
\vspace{-0.1in}
\end{table}

\begin{table}[!htb]
\vspace{-0.6em}
\setlength{\tabcolsep}{4pt}
\begin{center}
    \begin{tabular}{c|cc|cc}
    \toprule
        \multirow{2}{*}{\textbf{Output Structure}} & \multicolumn{2}{c|}{\textbf{HOTPOT}} & \multicolumn{2}{c|}{\textbf{2WIKI}} \\ 
         \cmidrule(lr){2-3} \cmidrule(lr){4-5} 
         & mean & std & mean & std \\
        \midrule
        \multicolumn{5}{c}{\textbf{Original Content}}  \\
        \midrule
        \multicolumn{1}{l|}{\textbf{Numbered Section Markdown Title}} & 0.5\% & 0.8\% & -0.5\% & 1.2\% \\ 
        \multicolumn{1}{l|}{\textbf{Numbered Section Original Title}} & 0.5\% & 0.8\% & -0.3\% & 1.2\% \\ 
        \multicolumn{1}{l|}{\textbf{Numbered Section Quote Title}} & 0.7\% & 1.1\% & -0.1\% & 1.5\% \\ 
        \multicolumn{1}{l|}{\textbf{Hierarchy Section Markdown Title}} & 0.5\% & 0.9\% & -0.4\% & 1.6\% \\ 
        \multicolumn{1}{l|}{\textbf{Hierarchy Section Quote Title}} & 0.7\% & 0.9\% & -0.3\% & 1.5\% \\ 
        \multicolumn{1}{l|}{\textbf{Star Section Markdown Title}} & 0.6\% & 0.7\% & -0.2\% & 1.4\% \\
        \multicolumn{1}{l|}{\textbf{Star Section Original Title}} & 0.7\% & 0.8\% & -0.1\% & 1.4\% \\
        \multicolumn{1}{l|}{\textbf{Star Section Quote Title}} & 0.7\% & 1.0\% & 0.2\% & 1.5\% \\
        \midrule
        \multicolumn{5}{c}{\textbf{Quote Content}}  \\
        \midrule
        \multicolumn{1}{l|}{\textbf{Numbered Section Markdown Title}} & 0.9\% & 1.2\% & 0.0\% & 1.5\% \\ 
        \multicolumn{1}{l|}{\textbf{Numbered Section Original Title}} & 1.0\% & 1.2\% & 0.1\% & 1.5\% \\ 
        \multicolumn{1}{l|}{\textbf{Numbered Section Quote Title}} & 0.9\% & 1.3\% & 0.5\% & 2.0\% \\
        \multicolumn{1}{l|}{\textbf{Hierarchy Section Markdown Title}} & 0.7\% & 1.2\% & -0.2\% & 1.8\% \\ 
        \multicolumn{1}{l|}{\textbf{Hierarchy Section Quote Title}} & 1.0\% & 1.4\% & 0.2\% & 1.7\% \\ 
        \multicolumn{1}{l|}{\textbf{Star Section Markdown Title}} & 0.8\% & 1.2\% & 0.2\% & 1.8\% \\
        \multicolumn{1}{l|}{\textbf{Star Section Original Title}} & 1.1\% & 1.2\% & 0.2\% & 1.7\% \\
        \multicolumn{1}{l|}{\textbf{Star Section Quote Title}} & 1.1\% & 1.5\% & 0.7\% & 2.1\% \\
    \bottomrule
    \end{tabular}
\end{center}\vspace{-0.6em}
\caption{Statistics of improved accuracy margin from various downstream LMs augmented by restructured outputs from \textit{Refiner}, compared with no structured output in multi-hop tasks.}
\label{tab:more-output-structure-multi-hop-result}
\vspace{-0.1in}
\end{table}

\clearpage
\newpage
\section{Case Study}
\label{sec: cases-in-single-hop-eval-dataset}
Below, we will provide several exemplary cases illustrating how our model assists downstream models in further extracting information to enhance accuracy.
% \subsection{Cases in Single-hop Evaluation Dataset}
We select one case from PopQA and two cases from TriviaQA subset under top 5 retrieval setting.
For case under PopQA, \textit{Refiner} successfully disseminates similar yet different information using section, improves the quality of answer of downstream LM. 
While in both of the cases under TriviaQA, the \textit{Refiner}-augmented models successfully answer with ground truths which are only mentioned indirectly.

\begin{table*}[ht!]
\scriptsize
\begin{tabular}{p{3cm}p{10cm}}
\toprule
\textbf{Question} & \textbf{What is George Lewis's occupation?} \\ 
\midrule
\textbf{Ground Truths} & \textbf{'politician', 'political leader', 'political figure', 'polit.', 'pol'} \\ 
\midrule

\textbf{Retrieved Document Chunks}
& 
\textbf{Title}: George Lewis (footballer, born 1913)\newline
\textbf{Content}: Thomas George Lewis (20 October 1913 – 6 August 1981) was a Welsh footballer who played as a centre-forward for Watford in the late 1930s, before joining Southampton for a brief period after World War II....Lewis spent the 1947–48 season back in the Third Division South with Brighton \& Hove Albion before joining Dartford, where he remained until 1952, when he took up a position as groundsman.
\newline
\newline
\textbf{Title}: George Lewis (footballer, born 2000)\newline
\textbf{Content}: George Lewis Igaba-Ishimwe Maniraguha (born 16 June 2000) is a professional footballer who plays as a forward for club Arsenal. Born in Kigali, Rwanda, Lewis moved to Tanzania at the age of one. He then moved to Norway the age of four, starting his career at local side Stakkevollan IF, before moving to Tromsdalen, and later Tromsø in 2015. He mostly played for the reserve teams of both clubs, notably notching 32 goals in 35 appearances over three seasons for Tromsdalen's reserve team.\newline
\newline
\textbf{Title}: George Robert Lewis\newline
\textbf{Content}: George Robert Lewis (1782–1871) was a versatile English painter of landscapes and portraits....Attribution \newline
\newline
\textbf{Title}: George E. Lewis\newline
\textbf{Content}: George Emanuel Lewis (born July 14, 1952) is an American composer, performer, and scholar of experimental music. He has been a member of the Association for the Advancement of Creative Musicians (AACM) since 1971, when he joined the organization at the age of 19. He is renowned for his work as an improvising trombonist and considered a pioneer of computer music, which he began pursuing in the late 1970s; in the 1980s he created Voyager, an improvising software he has used in interactive performances. Lewis's many honors include a MacArthur Fellowship and a Guggenheim Fellowship, and his book A Power Stronger Than Itself: The AACM and American Experimental Music received the American Book Award. Lewis is the Edwin H. Case Professor of American Music, Composition \& Historical Musicology at Columbia University.\newline
\newline

\textbf{Title}: George Lewis (rugby)\newline
\textbf{Content}: George Lewis (9 August 1897 – death unknown) was a Welsh rugby union and professional rugby league footballer who played in the 1920s and 1930s. He played club level rugby union (RU) for Pontypool RFC, as a centre, i.e. number 12 or 13, and representative level rugby league (RL) for Wales and Monmouthshire, and at club level for St. Helens, as a, i.e. number 1, 3 or 4, 6, or 7....George Lewis was the brother of the rugby union footballer for Pontypool RFC, and the rugby league for St. Helens; Stanley "Stan" Lewis.
 \\
\midrule 
\textit{Refiner} & {\color{green-color}1.1.} George Lewis (footballer, born 1913)\newline
Thomas George Lewis (20 October 1913 – 6 August 1981) was a Welsh footballer...\newline
{\color{green-color}1.2.} George Lewis (footballer, born 2000)\newline
George Lewis Igaba-Ishimwe Maniraguha (born 16 June 2000) is a professional footballer...\newline
{\color{green-color}2.1.} George E. Lewis\newline
George Emanuel Lewis (born July 14, 1952) is an American composer, performer, and scholar of experimental music. \\
\midrule
Llama2-7B-Chat & George Lewis's occupation is footballer. He played as a centre-forward for Watford in the late 1930s and joined Southampton for a brief period after World War II. \\
\midrule
\textit{Refiner} + Llama2-7B-Chat & Based on the content provided, there are three individuals with the name George Lewis:
\newline\newline
1. George Lewis (footballer, born 1913) - The content mentions that he was a Welsh footballer. Therefore, his occupation was a footballer.
\newline
2. George Lewis (footballer, born 2000) - The content mentions that he is a professional footballer. Therefore, his occupation is also a footballer.
\newline
3. George E. Lewis - The content does not mention his occupation directly, but based on the fact that he is a composer, performer, and scholar of experimental music, it can be inferred that his occupation is related to music, possibly as a composer or musician.
\newline\newline
Therefore, the answer to the question "What is George Lewis's occupation?" would be:
\newline\newline
* For George Lewis (footballer, born 1913) and George Lewis (footballer, born 2000): footballer.
\newline* For George E. Lewis: composer, performer, and scholar of experimental music." \\
\bottomrule
\end{tabular}
\caption{Case study of how \textit{Refiner} helps downstream LM to disseminate ambiguous information from PopQA evaluation set.}
\label{tab:popqa-example1}
\end{table*}

\begin{table*}[ht!]
\scriptsize
\begin{tabular}{p{3cm}p{10cm}}
\toprule
\textbf{Question} & \textbf{For which team did Babe Ruth blast his last Major League home run?} \\ 
\midrule
\textbf{Ground Truths} & \textbf{'boston braves', 'boston braves disambiguation', 'Boston Braves (disambiguation)', 'Boston Braves'} \\ 
\midrule
\textbf{Retrieved Document Chunks} & \textbf{Title}: Lou Chiozza\newline
\textbf{Content}: Braves, had been through for years but was still being exploited for his name. Just a few days before the Philadelphia series, the Babe hit three home runs in Pittsburgh. Chiozza, who had just been brought up from the Memphis Chicks, hit a short fly ball down the left field line that would have ordinarily at best been a double. Due to his advanced age and decreased mobility, Ruth stumbled after the ball in the outfield. The shortstop ran out, retrieved the ball and threw the ball home to barely stop Chiozza from an inside-the-park home run. Many in the park believed that the umpire felt sorry for the aging star and called an obviously safe Chiozza out at home plate. After Ruth realized that he was so slow that Lou almost made a home run on an ordinary base hit, he stood for a minute, folded his glove and walked off the field into the clubhouse. The Babe knew he was done and he officially retired a few days later on June 2, 1935. Chiozza recalled in his later life that he had wished Ruth had retired on a high note after hitting the three home runs in Pittsburgh \newline
\newline
\textbf{Title}: Guy Bush\newline
\textbf{Content}: \#714, and was a mammoth of a shot. It was the first home run to clear the right field grandstands at Forbes Field and was reportedly measured as a bomb. It was the last home run -- in fact, the last hit of any kind -- in Ruth's major league career, which ended less than a week later. Despite Ruth's goliath performance, Waite Hoyt, Ruth's former Yankee teammate, came in relief for Bush and won the game 11-7, albeit after initially blowing the save. The Pirates sent the 34-year-old Bush to the bullpen for the season. The transition did not \newline
\newline
\textbf{Title}: Babe Ruth\newline
\textbf{Content}: As part of the Yankees' vaunted "Murderers' Row" lineup of 1927, Ruth hit 60 home runs, which extended his MLB single-season record by a single home run. Ruth's last season with the Yankees was 1934; he retired from the game the following year, after a short stint with the Boston Braves. During his career, Ruth led the AL in home runs during a season twelve times. Ruth's legendary power and charismatic personality made him a larger-than-life figure during the Roaring Twenties. During his career, he was the target of intense press and public attention for his baseball exploits and off-field...a long fly caught by Paul Waner probably would have been a home run in any other ballpark besides Forbes Field. Ruth played in the third game of the Pittsburgh series on May 25, 1935, and added one more tale to his playing legend. Ruth went 4-for-4, including three home runs, though the Braves lost the game 11–7. The last two were off Ruth's old Cubs nemesis, Guy Bush. The final home run, both of the game and of Ruth's career, sailed out of the park over the right field upper deck–the first time anyone had hit a fair ball\newline
\newline
\textbf{Title}: Atlanta Braves\newline
\textbf{Content}: Brave before asking to be traded to the Milwaukee Brewers, while Ruth finished his career as a Boston Brave. In fact, until Barry Bonds eclipsed the 714 home runs hit by Babe Ruth in 2006, the top two home run hitters in Major League history had at one time been Braves. In 1976, the team was purchased by media magnate Ted Turner, owner of superstation WTBS, as a means to keep the team (and one of his main programming staples) in Atlanta. The financially strapped Turner used money already paid to the team for their broadcast rights as a down-payment.\newline
\newline
\textbf{Title}: 1935 Boston Braves season\newline
\textbf{Content}: the team due to the attention he commanded. He also discovered that rather than give him a share of the Braves' profits, Fuchs expected him to invest some of "his" money in the team. Seeing a team in utter collapse and realizing he was finished even as a part-time player, Ruth retired on June 1, six days after he had what remains one of the most memorable afternoons in baseball history. He clouted what turned out to be the last three home runs of his career in a game at Forbes Field while playing the Pittsburgh Pirates. He'd wanted to \\
\midrule 
\textit{Refiner} & 1.1. Babe Ruth
Ruth played in the third game of the Pittsburgh series on May 25, 1935, and added one more tale to his playing legend. Ruth went 4-for-4, including three home runs, though the Braves lost the game 11–7. The last two were off Ruth's old Cubs nemesis, Guy Bush. \newline
1.2. Atlanta Braves\newline
Brave before asking to be traded to the Milwaukee Brewers, while Ruth finished his career as a Boston Brave. \\ \midrule
Llama2-13B-Chat & According to the text, Babe Ruth's last Major League home run was hit off Guy Bush of the Pittsburgh Pirates at Forbes Field on May 25, 1935. \\
\midrule
\textit{Refiner} + Llama2-13B-Chat & Sure, I'd be happy to help! Based on the content provided, Babe Ruth played for the Boston Braves when he hit his last Major League home run. The content states that Ruth finished his career as a Boston Brave. Therefore, the answer to your question is the Boston Braves. \\
\bottomrule
\end{tabular}
\label{tab:triviaqa-example1}
\end{table*}

\begin{table*}[ht!]
\scriptsize
\begin{tabular}{p{3cm}p{10cm}}
\toprule
\textbf{Question} & \textbf{Apimania is an obsession with which creatures?} \\ 
\midrule
\textbf{Ground Truths} & \textbf{'Ground bees', 'bees', 'Solitary bee', 'Apiformes', 'Bee flight', 'ground bee', 'bee flight', 'anthophila', 'solitary bee', 'apiforme', 'Solitary bees', 'ground bees', 'Anthophila', 'Bee', 'Bees', 'apiformes', 'Apiforme', 'Ground bee', 'solitary bees', 'bee'} \\ 
\midrule
\textbf{Retrieved Document Chunks} & \textbf{Title}: Pedro Reyes (artist)\newline \textbf{Content}: Assembly of "pUN" took place in December 2015 at the 21st Century Museum of Contemporary Art in Kanazawa, Japan. Entomofagia (Spanish for entomophagy–the practice of eating insects) is a series of culinary experiments. The first happened in 2013 at Casa do Vidro in São Paulo, where a small snack cart offered visitors a Brazilian street food specialty: giant ants. The project focuses on the enormous ecological advantages of shifting protein consumption to insects. As an alternative to the ubiquitous fast-food staple, the beef hamburger, Reyes created the Grasswhopper, a burger with a patty made of crickets, a common snack in\newline\newline\textbf{Title}: Insects in literature\newline \textbf{Content}: Grasshopper" by Samson Samsonov (1955) and Jerry Paris (1970) based on that story. The Ancient Greek playwright Aeschylus has a gadfly pursue and torment Io, a maiden associated with the moon, watched constantly by the eyes of the herdsman Argus, associated with all the stars: "Io: Ah! Hah! Again the prick, the stab of gadfly-sting! O earth, earth, hide, the hollow shape—Argus—that evil thing—the hundred-eyed." William Shakespeare, inspired by Aeschylus, has Tom o'Bedlam in "King Lear", "Whom the foul fiend hath led through fire and through flame, through ford and whirlpool, o'er bog and quagmire", driven mad by the constant\newline \newline\textbf{Title}: Dromomania\newline \textbf{Content}: sometimes been clinical and pathologizing, and other times been descriptive of unusual enthusiasm without negative or medicalizing connotations, reflecting the diverse uses of the term mania itself. In the 17th century, that the term "mania" came to be used to describe any show of great enthusiasm for a specific activity or object. Later, it came to be used as a suffix for Greek words to refer to an irrational obsession, such as in the words "guanaikomania", "hippomania", and "nymphomania". At the same time emerged the French , such as in which was borrowed in English as "bibliomania". The original sense\newline---\newline\textbf{Title}: Erich Hoyt\newline \textbf{Content}: entomologist Ted Schultz) broke new ground. In Hoyt’s "delightful…multi-layered" "The Earth Dwellers", the "ant’s eye view of life works spectacularly" as Hoyt "fashions the ants into enchanting characters" charting "an insect’s course through sex, aggression and foreign policy". "Insect Lives", an American Library Association "Outstanding Book for the College Bound" is a "potpourri of fascinating excerpts written by some of the finest insect biologists and naturalists spanning many centuries." A deep sea book called "Creatures of the Deep" (Firefly, 2001) won the American Society of Journalists \& Authors, Inc. Outstanding Book Award, General Nonfiction. A second deep sea book, "Weird\newline\newline\textbf{Title}: Obsession (novel)\newline\textbf{Content}: Obsession (novel) Obsession is a mystery novel by American author Jonathan Kellerman. It is the 21st novel in the Alex Delaware series. Tanya Bigelow, a former patient, comes to Alex requesting help. She wants him and his friend Milo to investigate something her mother said on her death bed. Her mother told her that she did something terrible. No one believes that Tanya's mother Patty, who worked with Milo's partner at the hospital as a nurse, could have done anything terrible, but as the reader learns her past, it contains dark secrets. Alex has a new dog, Blanche, that Robin bought him after Spike died. Robin is living with Alex again. \\
\midrule
\textit{Refiner} & 1.1. Erich Hoyt
entomologist Ted Schultz) broke new ground. In Hoyt’s "delightful…multi-layered" "The Earth Dwellers", the "ant’s eye view of life works spectacularly" as Hoyt "fashions the ants into enchanting characters" charting "an insect’s course through sex, aggression and foreign policy".\newline
1.2. Insects in literature\newline
"Io: Ah! Hah! Again the prick, the stab of gadfly-sting! O earth, earth, hide, the hollow shape—Argus—that evil thing—the hundred-eyed."\newline
2.1. Obsession (novel)\newline
Her mother told her that she did something terrible. No one believes that Tanya's mother Patty, who worked with Milo's partner at the hospital as a nurse, could have done anything terrible, but as the reader learns her past, it contains dark secrets.\\
\midrule
Llama2-13B-Chat & The content provided does not directly mention an obsession with a specific creature related to the term "Apimania". However, it is possible to infer that the term "entomophagy" mentioned in the first paragraph refers to an interest or obsession with insects as a food source. The term "entomophagy" is derived from the Greek words "entomon" meaning insect and "phagein" meaning to eat. This practice has been explored in Pedro Reyes' art project "Entomofagia" which focuses on the advantages of shifting protein consumption to insects. Therefore, Apimania could refer to an obsession with insects. \\
\midrule
\textit{Refiner} + Llama2-13B-Chat & Based on the content provided, I can answer your question precisely.

According to the content, Apimania is an obsession with bees. This information can be inferred from the text "Obsession (novel)" and the quote "Io: Ah! Hah! Again the prick, the stab of gadfly-sting! O earth, earth, hide, the hollow shape—Argus—that evil thing—the hundred-eyed." which mentions bees and their sting.

Therefore, the answer to your question is: Bees. \\
\bottomrule
\end{tabular}
\caption{Two cases from TriviaQA evaluation set illustrating how \textit{Refiner} helps downstream LM to highlight the correct information.}
\label{tab:triviaqa-example2}
\end{table*}

%%%%%%%%%%%%%%%%%%%%%%%%%%%%%%%%%%%%%%%%%%%%%%%%%%%%%%%%%%%%

\end{document}